\documentclass[a4paper,12pt]{article}
\usepackage[margin=1in]{geometry}

\usepackage[utf8]{inputenc}

\usepackage[authoryear, round]{natbib}
\usepackage[symbols,nogroupskip]{glossaries-extra}
\usepackage{xspace}
\usepackage{algorithm}
\usepackage{algpseudocode}
\usepackage{subcaption}
\usepackage{siunitx}
\usepackage{booktabs}
\usepackage{hyphenat}
\usepackage{authblk}
\usepackage{amsmath}
\usepackage{amssymb}
\usepackage{amsthm}
\usepackage{graphicx}
\usepackage{hyperref}

\newcommand{\state}[1]{\ensuremath{\mathbf{s}_{#1}}\xspace}
\newcommand{\action}[1]{\ensuremath{\mathbf{a}_{#1}}\xspace}
\newcommand{\stateN}{\ensuremath{\mathbf{s}}\xspace}
\newcommand{\ob}[1]{\ensuremath{\mathbf{o}_{#1}}\xspace}
\newcommand{\obN}{\ensuremath{\mathbf{o}}\xspace}
\newcommand{\actionN}{\ensuremath{\mathbf{a}}\xspace}
\newcommand{\policy}[1]{\ensuremath{\pi(\mathbf{a}_{#1}|\mathbf{s}_{#1})}\xspace}
\newcommand{\dynamics}{\ensuremath{p(\state{t+1}|\state{t},\action{t})}\xspace}
\newcommand{\pinit}{\ensuremath{p_0(\mathbf{s})}\xspace}
\newcommand{\traj}{\ensuremath{\boldsymbol{\tau}}\xspace}
\newcommand{\logits}{\ensuremath{\nu(\mathbf{x})}\xspace}
\DeclareMathOperator*{\argmax}{arg\,max}
\renewcommand{\vec}[1]{\boldsymbol{#1}}

\newtheorem{theorem}{Theorem}
\newtheorem{lemma}{Lemma}
\newtheorem{corollary}{Corollary}
\newtheorem{proposition}{Proposition}

\glsxtrnewsymbol[description={a vector}]{x}{\ensuremath{\vec{x}}}
\glsxtrnewsymbol[description={a matrix}]{X}{\ensuremath{\vec{X}}}

\newacronym{AIL}{AIL}{adversarial imitation learning}
\newacronym{AIRL}{AIRL}{adversarial inverse reinforcement learning}
\newacronym{AICO}{AICO}{approximate inference control}
\newacronym{ARS}{ARS}{augmented random search}
\newacronym{BBVI}{BBVI}{black-box variational inference}
\newacronym{ELBO}{ELBO}{estimated lower bound objective}
\newacronym{EM}{EM}{expectation maximization}
\newacronym{ESS}{ESS}{elliptical slice sampling}
\newacronym{GAIL}{GAIL}{generative adversarial imitation learning}
\newacronym{NAIL}{NAIL}{non-adversarial imitation learning}
\newacronym{NES}{NES}{natural evolution strategies}
\newacronym{ONAIL}{O-NAIL}{offline non-adversarial imitation learning}
\newacronym{GAN}{GAN}{generative adversarial network}
\newacronym{GAIfO}{GAIfO}{generative adversarial imitation from observations}
\newacronym{GCL}{GCL}{guided cost learning}
\newacronym{GMM}{GMM}{Gaussian mixture model}
\newacronym{GVA}{GVA}{Gaussian variational approximation}
\newacronym{HFSGVI}{HFSGVI}{Hessian-free stochastic gradient variational inference}
\newacronym{HiREPS}{HiREPS}{hierarchical relative entropy policy search}
\newacronym{HMC}{HMC}{Hamiltonian Monte Carlo}
\newacronym{HVM}{HVM}{hierarchical variational models}
\newacronym{IAF}{IAF}{inverse autoregressive flows}
\newacronym{IL}{IL}{imitation learning}
\newacronym{iprojection}{I-Projection}{information projection}
\newacronym{IRL}{IRL}{inverse reinforcement learning}
\newacronym{IT}{IT}{inference tree}
\newacronym{KDE}{KDE}{kernel density estimation}
\newacronym{KL}{KL}{Kullback-Leibler (divergence)}
\newacronym{KLIEP}{KLIEP}{Kullback-Leibler importance estimation procedure}
\newacronym{LaDiPS}{LaDiPS}{layered direct policy search}
\newacronym{LBFGS}{L-BFGS}{limited-memory Broyden–Fletcher–Goldfarb–Shanno algorithm}
\newacronym{L-BFGS-B}{L-BFGS-B}{limited-memory Broyden–Fletcher–Goldfarb–Shanno algorithm, box-constrained}
\newacronym{LQR}{LQR}{linear quadratic regulator}
\newacronym{LQG}{LQG}{linear-quadratic-Gaussian control}
\newacronym{MaxCausalEntIRL}{MaxCausalEnt-IRL}{maximum causal entropy inverse reinforcement learning}
\newacronym{MaxEntIRL}{MaxEnt-IRL}{maximum entropy inverse reinforcement learning}
\newacronym{MCMC}{MCMC}{Markov chain Monte Carlo}
\newacronym{MDP}{MDP}{Markov decision process}
\newacronym{MLE}{MLE}{maximum likelihood estimation}
\newacronym{MMD}{MMD}{maximum mean discrepancy}
\newacronym{MORE}{MORE}{model-based relative entropy stochastic search}
\newacronym{MOTO}{MOTO}{model-free trajectory-based policy optimization}
\newacronym{mprojection}{M-Projection}{moment projection}
\newacronym{NPVI}{NPVI}{non-parametric variational inference}
\newacronym{NUTS}{NUTS}{the no u-turn sampler}
\newacronym{OLS}{OLS}{ordinary least squares}
\newacronym{PILCO}{PILCO}{probabilistic inference for learning control}
\newacronym{PL}{PL}{preference learning}
\newacronym{PTMCMC}{PTMCMC}{parallel tempering Markov chain Monte Carlo}
\newacronym{REPS}{REPS}{relative entropy policy search}
\newacronym{RKL}{RKL}{reverse Kullback-Leibler divergence}
\newacronym{RL}{RL}{reinforcement learning}
\newacronym{SAC}{SAC}{soft actor-critic}
\newacronym{SVGD}{SVGD}{Stein variational gradient descent}
\newacronym{VBOOST}{VBOOST}{variational boosting}
\newacronym{VI}{VI}{variational inference}
\newacronym{VIPS}{VIPS}{variational inference by policy search}

\begin{document}

\title{Non-Adversarial Imitation Learning and
its Connections to Adversarial Methods}

\author[1]{Oleg Arenz}
\author[2, 3]{Gerhard Neumann}
\affil[1]{Intelligent Autonomous Systems, TU Darmstadt}
\affil[2]{Autonomous Learning Robots, Karlsruhe Institute of Technology}
\affil[3]{Bosch Center for Artificial Intelligence, Renningen}

\date{}  

\maketitle

\begin{abstract}%
Many modern methods for imitation learning and inverse reinforcement learning, such as GAIL or AIRL, are based on an adversarial formulation. These methods apply GANs to match the expert's distribution over states and actions with the implicit state-action distribution induced by the agent's policy. However, by framing imitation learning as a saddle point problem, adversarial methods can suffer from unstable optimization, and convergence can only be shown for small policy updates. We address these problems by proposing a framework for non-adversarial imitation learning. The resulting algorithms are similar to their adversarial counterparts and, thus, provide insights for adversarial imitation learning methods. Most notably, we show that AIRL is an instance of our non-adversarial formulation, which enables us to greatly simplify its derivations and obtain stronger convergence guarantees. We also show that our non-adversarial formulation can be used to derive novel algorithms by presenting a method for offline imitation learning that is inspired by the recent ValueDice algorithm, but does not rely on small policy updates for convergence. In our simulated robot experiments, our offline method for non-adversarial imitation learning seems to perform best when using many updates for policy and discriminator at each iteration and outperforms behavioral cloning and ValueDice.
\end{abstract}

\newpage

\section{Introduction}
Imitation learning \citep[\gls{IL},][]{Schaal1999, osa2018} and inverse reinforcement learning \citep[\gls{IRL},][]{Ng2000} are two related areas of research that aim to teach agents by providing demonstrations of the desired behavior. Whereas imitation learning aims to learn a \emph{policy} that results in a similar behavior, inverse reinforcement learning focuses on inferring a \emph{reward function} that might have been optimized by the demonstrator, aiming to better generalize to different environments. Both areas of research are often formalized as distribution-matching, that is, the learned policy (or the optimal policy for \gls{IRL}) should induce a distribution over states and actions that is close to the expert's distribution with respect to a given (usually non-metric) distance. Commonly applied distances are the forward Kullback-Leibler (\gls{KL}) divergence~\citep[e.g., ][]{Ziebart2010a}, which maximizes the likelihood of the demonstrated state-action pairs under the agent's distribution, and the reverse Kullback-Leibler (\gls{RKL}) divergence~\citep[e.g.,][]{Arenz2016,Fu2018,Ghasemipour2020} which minimizes the expected discrimination information~\citep{Kullback1951} of state-action pairs sampled from the agent's distribution.  
However, since the emergence of generative adversarial networks \citep[\glspl{GAN},][]{Goodfellow2014} as a solution technique for both areas, other divergences have been investigated such as the Jensen-Shannon divergence~\citep{Ho2016}, the Wasserstein distance~\citep{Xiao2019} and general $f$-divergences~\citep{Ke2019, Ghasemipour2020}. Although \glspl{GAN} are typically not applied to time-series data, their application for imitation learning is surprisingly straightforward. The discriminator can typically be trained in the exact same way---agnostically to the data-generation process---by aiming to discriminate state-action samples of the agent and the expert. The generator objective, on the other hand, can be typically solved using an off-the-shelf reinforcement learning \citep[\gls{RL},][]{SuttonBarto1998} algorithm. Apart from this flexibility, the adversarial formulation is also appealing for scaling to neural network policies and reward functions and for its efficiency, since unlike some previous approaches~\citep[e.g.,][]{Ratliff2006, Ziebart2010a}, they do not require to solve a full reinforcement learning problem iteratively but only perform few policy updates at each iteration. However, it is often difficult to achieve stable optimization with adversarial approaches. Firstly, the discriminator typically relies on an estimate of the probability density ratio which is difficult to approximate for high-dimensional problems, especially for those areas of the state-action-space that are not encountered by the current policy or the expert. Secondly, the reward signal provided by the discriminator is specific to the current policy and, thus, can quickly become invalid if the generator is updated to greedily. 

In this work, we directly address the latter problem. We derive an upper bound on the reverse Kullback-Leibler divergence between the agent's and expert's distribution which allows us to guarantee improvement even for large policy updates (when assuming an optimal discriminator). By iteratively tightening and optimizing this bound similar to expectation-maximization, we can show convergence to the optimal solution. 
Similar to adversarial methods, the reward signal in our non-adversarial formulation is based on a density-ratio that is learned by training a discriminator to classify samples from the agent's distribution and the expert's distribution. However, in contrast to the adversarial formulation, our reward function is explicitly defined with respect to the density ratio based on the agent's previous distribution which is achieved by adding an additional term that penalizes the divergence to the last policy. However, our non-adversarial formulation is not only closely connected to adversarial imitation learning, but also to inverse reinforcement learning: As our reward signal is not specific to the current policy, it can serve as reward function for which any maximizing policy matches the expert demonstrations.

Indeed, we show that adversarial inverse reinforcement learning \citep[\gls{AIRL},][]{Fu2018} learns and optimizes the lower bound reward function of our non-adversarial formulation suggesting that \gls{AIRL} is better viewed as a non-adversarial method. To the best of our knowledge, the theoretical justification of \gls{AIRL} is currently not well understood. For example, \citet{Fu2018} justify \gls{AIRL} as an instance of maximum causal entropy inverse reinforcement learning \citep[\gls{MaxCausalEntIRL},][]{Ziebart2010a} by relating the update of their reward function---which corresponds to an energy-based model of the policy---with the gradient of the maximum likelihood (forward KL) objective of \gls{MaxCausalEntIRL}. However, we clarify that the gradients of the different objectives only coincide after convergence and, furthermore, only when the expert demonstrations are perfectly matched and, thus, any divergence is minimized. Furthermore, \gls{AIRL} is not a typical adversarial method, since the discriminator directly depends on the policy and is, thus, not held constant during the generator update. To the best of our knowledge, it has never been investigated whether the theoretical analyses of generative adversarial nets apply to this setting. However, based on our non-adversarial formulation, we show that \gls{AIRL} indeed enjoys stronger convergence guarantees than adversarial methods since we can drop the requirement of sufficiently small policy updates at each iteration and instead only assume improvement with respect to the current reward function.

Apart from deepening our understanding of \gls{AIRL}, our non-adversarial formulation gives rise to novel algorithms for imitation learning and inverse reinforcement learning. For example, we show that the Q-function for our non-adversarial reward function can be estimated offline based on DualDice~\citep{Nachum2019}, a method for offline density-ratio estimation. The resulting algorithm is closely connected to the recently proposed imitation learning method ValueDice~\citep{Kostrikov2020} but does not involve solving a saddle point problem. 

The contributions of this work include 
\begin{itemize}
	\item introducing non-adversarial imitation learning (\gls{NAIL}), a framework for imitation learning that resembles adversarial imitation learning but enjoys stronger convergence guarantees by not involving a saddle point problem,
	\item presenting an alternate derivation for adversarial inverse reinforcement learning based on our non-adversarial formulation,
	\item introducing offline non-adversarial imitation learning (\gls{ONAIL}), an instance of non-adversarial imitation learning that does not involve interactions with the environment, and
	\item presenting a more general derivation of several adversarial imitation learning methods \citep{Ho2016, Torabi2018, Ghasemipour2020} that is based on matching noisy trajectory observations.
\end{itemize}

The remainder of this article is structured as follows. We formally specify our more general formulation for imitation learning and discuss adversarial imitation learning and \gls{AIRL} in Section~\ref{nail_sec:preliminaries}. Our derivations for non-adversarial imitation learning are presented in Section~\ref{nail_sec:nail}. In Section~\ref{nail_sec:AirlAsNail}, we investigate \gls{AIRL} through the lens of non-adversarial imitation learning. In Section~\ref{nail_sec:onail}, we present an offline imitation learning algorithm based on our non-adversarial formulation, and in Section~\ref{nail_sec:experiments} we present experimental results. The main insights from our work are discussed in Section~\ref{nail_sec:discussion}.

\section{Preliminaries}
\label{nail_sec:preliminaries}
After formalizing the problem setting, we will discuss how \glspl{GAN} can be applied for imitation learning. Furthermore, we will discuss the modifications employed by \gls{AIRL} for extracting a reward function.

\subsection{Problem Formulation}
We consider a Markov decision process for the discounted, infinite horizon setting. At each time step $t$, an agent observes the state \state{t} and uses a stochastic policy \policy{t} to choose an action \action{t}. Afterwards, with probability $\gamma < 1$, the agent transitions to the next state \state{t+1} according to stochastic system dynamics \dynamics, which we do not assume to be known. With probability $1 - \gamma$, the episode ends and the environment gets reset to an initial state \state{0} drawn from the initial state distribution \pinit. By assuming such environment resets, we introduce a discounting of future rewards---which is commonly applied in practice---and ensure existence of a stationary distribution, which includes transient behavior~\citep{Hoof2017}.
We refer to the tuple containing the states and actions encountered during an episode as trajectory $\traj_i=(\state{0}, \action{0}, \dots, \state{T_i}, \action{T_i})$, where \state{T_i} and \action{T_i} correspond to the last state and action encountered at episode $i$. A given policy $\pi$ induces a distribution over trajectories $p^\pi(\traj)$ and for each time step $t$ a distribution over states and actions $p^\pi_t(\stateN, \actionN)$ (which can be computed from $p^\pi(\traj)$ by marginalization). The stationary distribution over states and actions induced by a policy $\pi$ is given by $p^\pi(\stateN, \actionN) = (1-\gamma) \sum_{t=0}^\infty \gamma^t p_t^\pi(\mathbf{s}, \mathbf{a})$. 

In a reinforcement learning setting, the agent would further obtain a reward  $r(\state{t}, \action{t})$ at each time step $t$ and would aim to find a policy that maximizes the expected reward $J_\text{RL} = \int_{\stateN,\actionN} p^\pi(\stateN, \actionN) r(\stateN, \actionN) d\stateN d\actionN$. 
However, for the purpose of this work, we do not assume that a reward function is available. Instead, we consider the problem of imitation learning from observations, which generalizes state-action based imitation learning. Namely, we assume that the agent observes a set of $N$ expert demonstrations $\mathcal{D}=\left\{\mathbf{o}_i\right\}_{1\le i \le N}$ in some observation space $\mathcal{O}$.
We further assume that $p(\mathbf{o}|\boldsymbol{\tau})$, the probabilistic mapping from the agent's trajectory to a distribution of observations, is given. For example, if the observation space is given by the end-effector pose of a robot and the state-space includes the robot's joint position, the mapping from trajectory to observation space would be given as the distribution over end-effector poses during the trajectory, which can be computed from the states using the robot's forward kinematics. We do not assume that the states and actions of the expert are observed nor do we assume that the expert acts in the same \gls{MDP} as the agent. In the aforementioned example, the demonstrations might be recorded by tracking the hand pose of a human expert. Furthermore, depending on the probabilistic mapping, we can match distributions over full trajectories, individual steps or transitions. This formulation generalizes the setting of several imitation learning methods such as, \gls{GAN}-\gls{GCL}~\citep{Finn2016a}, \gls{GAIL}~\citep{Ho2016}, \gls{AIRL}~\citep{Fu2018} or \gls{GAIfO}~\citep{Torabi2018}.
In imitation learning, we aim to minimize a given divergence $D(p^\pi(\obN)||q(\obN))$ between the agent's distribution over observations $p^\pi(\obN)$ and the expert's distribution over observations $q(\obN)$, that is,
\begin{equation}
\label{nail_eq:generalILobjective}
J_\text{IL} = \min_\pi D(p^\pi(\obN)||q(\obN)).
\end{equation}
Optimizing Objective~\ref{nail_eq:generalILobjective} is complicated by the fact that the agent's distribution over observations $p^\pi(\obN)$ is not analytically known and can only be controlled implicitly by changing the agent's policy.

\subsection{Generative Adversarial Nets}
We will now briefly review (Jensen-Shannon-)\glspl{GAN} and their generalization to general $f$-divergences. We will also discuss the connection between density-ratio estimation and optimizing a discriminator.

\subsubsection{Jensen-Shannon-GANs}
Generative adversarial nets~\citep{Goodfellow2014} are a popular technique to train implicit distributions to produce samples that are similar to samples from an unknown target distribution. Implicit distributions are distributions for which the density function is implicitly defined by a sampling procedure but not explicitly modeled. \citet{Goodfellow2014} proposed to minimize the Jensen-Shannon Divergence $D_\text{JS}(p(\mathbf{x}||q(\mathbf{x}))$ between an implicit distribution $p(\mathbf{x})$ and a data distribution $q(\mathbf{x})$ by solving the saddle point problem
\begin{equation}
\label{nail_eq:GAN}
J_\text{GAN}(p, D) = \min_p \max_D \mathrm{E}_{\mathbf{x} \sim q}\left[ \log D(\mathbf{x}) \right] + \mathrm{E}_{\mathbf{x} \sim p(\mathbf{x})}\left[ \log\left(1 - D(\mathbf{x})\right) \right].
\end{equation}
Here, $D(\mathbf{x})$ (typically a neural network) is called \emph{discriminator} and assigns a scalar value in the range $]0,1[$ to each sample $\mathbf{x}$. The \emph{generator} $p(\mathbf{x})$ is typically represented as a neural network $G(\mathbf{z})$ that transforms Gaussian input noise $\mathbf{z} \sim \mathcal{N}(\mathbf{z})$. However, the derivations provided by \citet{Goodfellow2014} also apply for general implicit distributions $p(\mathbf{x})$.
\citet{Goodfellow2014} show that solving the saddle point problem (Eq.~\ref{nail_eq:GAN}) minimizes the Jensen-Shannon divergence and that the optimal solution can be found by alternating between optimizing the discriminator to convergence and performing small generator updates. However, in practice, the discriminator also obtains only slight updates at each iteration in order to avoid vanishing gradients~\citep{Arjovsky2017}. The discriminator objective corresponds to minimizing the binary cross-entropy loss which is commonly used for binary classification. Hence, the discriminator is trained to classify samples from the data distribution and samples from the generator. The generator objective does not involve the probability density of the generator and can, thus, also be optimized for implicit distributions. Whereas, the original formulation by~\citet{Goodfellow2014} minimizes the Jensen-Shannon Divergence, similar ideas can also be applied to other divergences~\citep{Arjovsky2017a, Nowozin2016}. We will briefly review $f$-\glspl{GAN}~\citep{Nowozin2016}, which can minimize general ${f}$-divergences because the family of ${f}$-divergences also includes the reverse Kullback-Leibler divergence which is central to this work.

\subsubsection{$f$-GANs}
The family of ${f}$-divergences is defined for a given convex, lower-semicontinuous function $f$ as
\begin{equation*}
D_f(q(\mathbf{x})||p(\mathbf{x})) = \mathrm{E}_p\left[f\Big(\frac{q(\mathbf{x})}{p(\mathbf{x})}\Big)\right].
\end{equation*}
\citet{Nowozin2016} showed that we can minimize the $f$-divergence between the data distribution $q$ and an implicit distribution $p$ by solving the saddle point problem
\begin{equation}
\label{nail_eq:f-gan}
J_\text{F-GAN}(p, D) = \min_p \max_D \mathrm{E}_{\mathbf{x} \sim q}\left[  D(\mathbf{x}) \right] - \mathrm{E}_{\mathbf{x} \sim p(\mathbf{x})}\left[ f^*(D(\mathbf{x})) \right],
\end{equation}
where $$f^*(t)=\sup_{u \in \text{dom}_f} ut - f(u)$$ is the convex conjugate of $f$. 
Depending on the choice of $f$, we need to ensure that the output of the discriminator respects the domain of the convex conjugate, that is $\forall_\mathbf{x}: D(\mathbf{x}) \in \text{dom}_{f^*}$, which can be achieved by applying an appropriate output activation. \citet{Nowozin2016} provide for several common choices of $f$ their respective convex conjugates and suitable output activations. \citet{Nowozin2016} also note, that the optimal discriminator $D^\star(\mathbf{x})$ for a given generator $p(\mathbf{x})$ corresponds to the derivative of $f$ evaluated at the density-ratio $\frac{q(\mathbf{x})}{p(\mathbf{x})}$, that is, $$D^\star(\mathbf{x}) = f'\Big(\frac{q(\mathbf{x})}{p(\mathbf{x})}\Big).$$ Hence, for any $f$-divergence and when assuming the discriminator to be optimal, optimizing the $f$-\gls{GAN} objective given by Equation~\ref{nail_eq:f-gan} with respect to the generator, 
\begin{equation}
\underset{p}{\max}\; \mathrm{E}_{\mathbf{x} \sim p(\mathbf{x})}\left[ f^*(D^\star(\mathbf{x})) \right] =  \underset{p}{\max}\; \mathrm{E}_{\mathbf{x} \sim p(\mathbf{x})}\left[ f^* \circ f'\Big(\frac{q(\mathbf{x})}{p(\mathbf{x})}\Big) \right] =  \underset{p}{\max} \; \mathrm{E}_{\mathbf{x} \sim p(\mathbf{x})}\left[ g\Big(\frac{q(\mathbf{x})}{p(\mathbf{x})}\Big) \right],
\end{equation}
corresponds to maximizing the expected value of a function $g = f^\star \circ f'$ of the density-ratio $\phi(\mathbf{x})=\frac{q(\mathbf{x})}{p(\mathbf{x})}$, which closely connects the problem of minimizing $f$-divergences with density-ratio estimation.

\subsubsection{A Connection to Density-Ratio Estimation}
\label{nail_sec:dre}
Density-ratio estimation considers the problem of estimating the density-ratio $\phi(\mathbf{x})$ based on samples from $q(\mathbf{x})$ and $p(\mathbf{x})$. A naive approach would perform density estimation to independently estimate both distributions and then approximate the density ratio based on these estimates. However, while it is possible to compute the density ratio from the individual density, it is not possible to recover the individual densities from their ratio, and, thus, density-ratio estimation is a simpler problem than density estimation~\citep{Sugiyama2012}. 
Indeed, density-ratio estimation is closely related to binary classification~\citep{Menon2016}. 

To illustrate this connection, consider the following binary classification task. Assume that we aim to classify samples that have been drawn with probability $z(Q)$ from the distribution $q(\mathbf{x})$ and with probability $z(\neg Q)=1-z(Q)$ from the distribution $p(\mathbf{x})$. Hence, we consider the mixture model 
\begin{align*}
z(\mathbf{x}) = z(Q) q(\mathbf{x}) + (1 - z(Q)) p(\mathbf{x}),
\end{align*}
where the class frequencies $z(Q)$ and $z(\neg Q)$ are known, and we want to learn a model $\tilde{z}(Q|\mathbf{x})$ to approximate the conditional class probabilities $z(Q|\mathbf{x})$---for example, by minimizing the expected cross entropy between $z(Q|\mathbf{x})$ and $\tilde{z}(Q|\mathbf{x})$ as in the inner maximization in Eq.~\ref{nail_eq:GAN}. The model is typically represented by squashing learned logits \logits with a sigmoid, that is
\begin{equation*}
\tilde{z}(Q|\mathbf{x}) = \frac{1}{1+\exp{\left(-\logits\right)}} \Rightarrow \logits = \log \Big(\frac{\tilde{z}(Q|\mathbf{x})}{1 - \tilde{z}(Q|\mathbf{x})}\Big).
\end{equation*}
At the optimum, where $\tilde{z}(Q|\mathbf{x}) \approx z(Q|\mathbf{x})$, we have
\begin{equation}
\label{nail_eq:dre_classification}
\logits \approx \log \Big( \frac{z(Q|\mathbf{x})}{1-z(Q|\mathbf{x})} \Big) = \log \Big( \frac{z(Q) q(\mathbf{x})}{z(\neg{Q}) p(\mathbf{x})} \Big) = \log \Big( \frac{z(Q)}{z(\neg{Q})} \Big) + \log \Big( \phi(\mathbf{x}) \Big).
\end{equation}
Hence, when choosing equal class frequencies, that is $z(Q)=z(\neg Q)=0.5$, the logits \logits of the optimal classifier approximate the log density-ratio $\log(\phi(\mathbf{x}))$. 

\subsection{Adversarial Imitation Learning}
As generative adversarial nets can be used to minimize a variety of different divergences between an implicit distribution and the data distribution, they are suitable also for imitation learning. When applying \glspl{GAN} to imitation learning, the generator is given by the distribution over observations induced by the agent's policy, $p^\pi(\obN) = \int_{\traj} p^\pi(\traj) p(\obN|\traj) d\traj$. \citet{Ho2016} introduced this idea when they proposed generative adversarial imitation learning (\gls{GAIL}), where they applied the Jensen-Shannon-\gls{GAN} objective (Eq.~\ref{nail_eq:GAN}). However, we will directly consider the more general $f$-\gls{GAN} objective given by Equation~\ref{nail_eq:f-gan}, which was congruently proposed for imitation learning by \citet{Ke2019} and \cite{Ghasemipour2020}. 

\subsubsection{$f$-GANs for Imitation Learning}
When using an $f$-\gls{GAN} for imitation learning, the saddle-point problem is given by
\begin{equation}
\label{nail_eq:f-gail}
J_\text{AIL}(\pi, D) = \min_\pi \max_D \mathrm{E}_{\obN \sim q}\left[  D(\obN) \right] - \mathrm{E}_{\obN \sim p^\pi(\obN)}\left[ f^*(D(\obN)) \right].
\end{equation}
Please note that the minimization is still performed over $p^\pi(\obN)$ which we can only affect through $\pi$. 
Optimizing the discriminator is performed in the same way as for standard $f$-\glspl{GAN} using samples from $p^\pi(\obN)$ that can be obtained by executing the policy $\pi$. In the trajectory-centric formulation, where we make no further assumptions than $p^\pi(\obN) = \int_{\traj} p^\pi(\traj) p(\obN|\traj) d\traj$, optimizing the generator corresponds to an episodic reinforcement learning problem
\begin{equation}
\label{nail_eq:episodic_rl}
\max_\pi \mathrm{E}_{\traj \sim p^\pi(\traj)}\left[ r_\text{ep}(\traj) \right],
\end{equation}
for the episodic reward
\begin{equation*}
r_\text{ep}(\traj) = \int_\obN p(\obN|\traj) f^*(D(\obN)) d\obN.
\end{equation*}
As we can only obtain a reward signal for full trajectories, we can not apply standard reinforcement learning algorithms that typically assume Markovian rewards $r(\stateN, \actionN)$ or $r(\stateN, \actionN, \stateN')$, where $\stateN'$ is the state at the next time step. Instead, the policy can be optimized using stochastic search or black-box optimizers ~\citep{Abdolmaleki2015, Hansen2003}.

Hence, adversarial imitation learning methods are typically restricted to step-based distribution matching. Furthermore, they often assume direct observations of states and actions, that is, they aim to match $q(\stateN, \actionN)$~\citep{Ho2016, Xiao2019}, $q(\stateN)$~\citep{Ghasemipour2020}, or $q(\stateN, \stateN')$~\citep{Torabi2018}. 
We can incorporate such restrictions by making further assumptions on the form of $p(\obN|\traj)$ as shown by Proposition~\ref{nail_prop:stateReward}. 

\begin{proposition}
	\label{nail_prop:stateReward}
	Assume that the distribution over observations at a given time step $t$ is completely characterized by the state $\state{t}$ and action $\action{t}$ at that time step, and hence,
	\begin{equation*}
	p(\obN|\traj, t) = p(\obN| \state{t}^{\traj}, \action{t}^{\traj}).
	\end{equation*}
	Then, the episodic reinforcement learning problem given by Equation~\ref{nail_eq:episodic_rl} can be solved by maximizing the Markovian reward function
	\begin{equation}
	\label{nail_eq:adv_reward}
	r_{\text{adv}}(\stateN, \actionN) = \int_\obN p(\obN|\stateN, \actionN) f^*(D(\obN)) d\ob.
	\end{equation}
	\begin{proof}
		See Appendix~\ref{nail_app:proof_markovian_reward}. Intuitively, when the distribution over observations can be computed independently for each time step using $p(\obN|\stateN, \actionN)$, we can obtain $p(\obN)$ also by marginalizing based on the stationary distribution $p^\pi(\stateN, \actionN)$, which turns the optimization over the policy in Eq.~\ref{nail_eq:f-gail} into a step-based reinforcement learning problem.
	\end{proof}
\end{proposition}

While the restriction to step-based distribution matching is necessary for obtaining Markovian rewards, assuming direct observerations of states and action is in general not necessary. Indeed, dropping this assumption greatly generalizes the problem formulation of imitation learning, for example, by enabling us to imitate certain features (e.g. hand poses) of a human demonstration without assuming that their states and actions can be observed or mapped to the agent's MDP. However, while our derivations for non-adversarial and adversarial imitation learning are formulated for general observations, both instances of \gls{NAIL} that we will discuss in Section~\ref{nail_sec:instances} assume direct state-action observations which are exploited by the optimization.

Proposition~\ref{nail_prop:stateReward} can be straightforwardly extended to also include the next state by assuming that $p(\obN|\traj, t) = p(\obN| \state{t}^{\traj}, \action{t}^{\traj}, \state{t+1}^{\traj})$, and when maximizing the reward with respect to the stationary distribution $p^{\pi}(\stateN, \actionN, \stateN')$. However, the observation model needs to treat the last time step of an episode $\traj_i$ differently, because we can not make use of the next state $\stateN'$ which corresponds to the first time step of the next episode $\traj_{i+1}$ and thus can not affect $p(\obN|\traj_{i})$. Instead, we could emit a distinct observation $\obN_{reset}$---for example, by setting a special flag~\citep{Kostrikov2018}---for final time steps. However, in the following, we will only consider the more common imitation learning setting, where the observations and, thus, the rewards at time step $t$ only depend on $\state{t}$ and $\action{t}$.



\subsubsection{Choosing a Divergence}
\label{nail_sec:choosing_f}
The exact form of the reward function depends on the choice of $f$ and, thus, the divergence that we want to minimize. When assuming that the expert's distribution can be matched exactly, all divergences yield the same optimal solution $p^\pi(\obN) = q(\obN)$. However, especially due to restriction on the agent's policy---which is often Gaussian in the actions for a given state---matching the expert's distribution exactly is often not feasible. In such cases, the choice of divergence has large impact on the solution. For example, it is well-known that the forward \gls{KL} divergence
\begin{equation*}
D_\text{KL}(q(\obN)||p^\pi(\obN)) = D_f^{\text{KL}}(p^\pi(\obN)||q(\obN)) = \mathrm{E}_q\left[\log\Big(\frac{q(\obN)}{p^\pi(\obN)}\Big)\right],
\end{equation*}
which corresponds to $f(u)=u \log(u)$, results in a mode averaging behavior. The agent's policy will maximize the likelihood of all expert demonstrations, which may require putting most of its probability mass on areas where we do not have expert observations. As the resulting behavior can potentially be dangerous, it is often argued that the reverse \gls{KL} divergence should be preferred for imitation learning~\citep{Finn2016a, Ghasemipour2020}.
The reverse \gls{KL} divergence, 
\begin{equation*}
D_\text{KL}(p^\pi(\obN)||q(\obN)) = D_{f}^{\text{RKL}}(p^\pi(\obN)||q(\obN)) = \mathrm{E}_{p^\pi}\left[\log\Big(\frac{p^\pi(\obN)}{q(\obN)}\Big)\right],
\end{equation*}
is obtained when choosing $f(u) = -\log(u)$ and typically results in a mode-seeking behavior. The resulting policy will assign high likelihood to as many expert observations as possible while avoiding putting much probability mass on areas where we do not have expert observations. Although the resulting behavior might not exhibit the same variety as the expert, it will avoid producing any observations that are significantly different from the expert demonstrations, resulting in a safer behavior. Hence, we also focus on the reverse \gls{KL} divergence for deriving our non-adversarial imitation learning formulation. 
The convex conjugate and the derivative for $f_\text{RKL}(u) = -\log(u)$ are given by
\begin{align*}
f^*_\text{RKL}(t) = -1 - \log(-t)  \quad \quad \text{and} \quad \quad  f'_\text{RKL}(u) = - \frac{1}{u}.
\end{align*}
For an optimal discriminator, the observation-based reward function is, thus,
$$
r(\obN) = f^*_\text{RKL} \circ f'_\text{RKL}\big(\frac{q(\obN)}{p^\pi(\obN)}\big) = -1 + \log\Big(\frac{q(\obN)}{p^\pi(\obN)}\Big)
$$
or 
\begin{equation}
\label{nail_eq:adversarial_reward}
r_{adv}(\stateN, \actionN) = -1 + \log\Big(\frac{q(\stateN,\actionN)}{p^\pi(\stateN, \actionN)}\Big)
\end{equation}
when we perform step-based matching and directly observe state and actions. 
As the domain of $f^{*}_\text{RKL}$ is $\mathbb{R}^{-}$, a suitable output activation for the discriminator is $D(\obN) = -\exp(\nu(\obN))$ \citep{Nowozin2016}.
The discriminator loss based on Equation~\ref{nail_eq:f-gan} for the reverse \gls{KL}-divergence is then given by
\begin{equation}
\label{nail_eq:fgan_rkl_objective}
J_{\text{RKL-FGAN}}(\nu) = \underset{\nu}{\max} \; \mathrm{E}_{\obN \sim p^{\pi}(\obN)}\left[ \nu(\obN) \right] -\mathrm{E}_{\obN \sim q(\obN)}\left[  \exp{\nu(\obN)} \right] + 1,
\end{equation}
where the optimal solution $$\nu^\star(\obN) = \log \left(- D^\star(\obN)\right) = \log \left( -f'_{\text{RKL}}\Big(\frac{q(\obN)}{p^{\pi}(\obN)}\Big)\right) = \log \Big(\frac{p^{\pi}(\obN)}{q(\obN)}\Big) = -r(\obN) - 1$$ corresponds to the negated log density-ratio and thus to the cost function $-r(\obN)$, apart from a constant offset that does not affect the generator optimization. Indeed, the $f$-\gls{GAN} discriminator objective for the reverse \gls{KL} is also known as a generalized form~\citep{Menon2016} of Kullback-Leibler importance estimation (\gls{KLIEP})~\citep{Sugiyama2008}, which is well-known loss for density-ratio estimation.
Instead of learning the density-ratio based on Equation~\ref{nail_eq:fgan_rkl_objective}, we can also use the discriminator objective of the Jensen-Shannon-\gls{GAN}~(Eq.~\ref{nail_eq:GAN}), where---as shown in Section~\ref{nail_sec:dre}---the logits of the discriminator also converge to the log density-ratio. 

We will now review adversarial inverse reinforcement learning (\gls{AIRL}) which also minimizes the reverse \gls{KL}, albeit it uses a reward function that subtly differs from the adversarial formulation given by Equation~\ref{nail_eq:adversarial_reward}.

\subsection{Adversarial Inverse Reinforcement Learning}
Adversarial inverse reinforcement learning was derived as an instance of maximum causal entropy inverse reinforcement learning (MaxEnt-IRL, \citep{Ziebart2008}), which minimizes the \emph{forward} KL to the expert distribution. Hence, we will briefly discuss MaxEnt-IRL.

\subsubsection{Interlude: Maximum Causal Entropy IRL}
\gls{MaxCausalEntIRL} assumes the expert to use the policy 
\begin{equation}
\label{nail_eq:maxent_expert_model}
\pi_\text{expert}(\actionN| \stateN) = \exp{\Big({Q^\text{soft}(\stateN, \actionN)}-{V^\text{soft}(\stateN)}\Big)} = \exp{\big(A^\text{soft}\big)}.
\end{equation}
Here, the soft-Q function $Q^\text{soft}$, the soft-value function $V^\text{soft}$ and the soft-advantage function $A^\text{soft}$ are defined as~\citep{Ziebart2010a}
\begin{equation}
\begin{aligned}
Q^\text{soft}(\stateN, \actionN) &= r_\text{expert}(\stateN, \actionN) + \gamma \int_{\stateN'} p(\stateN'|\stateN, \actionN) V^\text{soft}(\stateN') d\stateN', \\
V^\text{soft}(\stateN) &= \log \int_\actionN \exp\big(Q^\text{soft}(\stateN, \actionN)\big) 
d\actionN, \\
A^\text{soft}(\stateN, \actionN) &= Q^\text{soft}(\stateN, \actionN) - V^\text{soft}(\stateN).
\end{aligned}
\end{equation}
The soft-value function $V^\text{soft}(\stateN)$ is the log-normalizer of $\pi$, but also corresponds to the value (which includes the expected entropy to come) of state $\stateN$ for the optimal policy of the entropy-regularized reinforcement learning problem
\begin{equation}
\label{nail_eq:maxent_rl}
\underset{\pi}{\max} \; J_\text{maxent-RL}(\pi) = \underset{\pi}{\max} \int_{\stateN,\actionN} p^\pi(\stateN,\actionN) \left( r_\text{expert}(\stateN, \actionN) - \log \pi(\actionN|\stateN) \right) d\stateN d\actionN.
\end{equation}
The expert model (Equation~\ref{nail_eq:maxent_expert_model}) is indeed well-motivated as it follows the principle of maximum (causal) entropy~\citep{Jaynes1957, Ziebart2010a}. Namely, \citet{Ziebart2010a} showed that this expert model corresponds to the maximum entropy policy that matches given empirical features $\tilde{\mathbf{f}}(\stateN, \actionN)$ in expectation for a linear reward function $$r_\text{expert}(\stateN, \actionN) = \boldsymbol{\theta}^\top \mathbf{f}(\stateN, \actionN ),$$ where the feature function $\mathbf{f}$ is assumed to be given.

Maximum causal entropy \gls{IRL} learns the parameters $\boldsymbol{\theta}$ of the expert's reward function by maximizing the likelihood of the expert demonstrations
\begin{equation}
\label{nail_eq:maxentLL}
\begin{aligned}
\mathcal{L} &= \mathrm{E}_{\stateN,\actionN \sim q(\stateN,\actionN)} \left[ \left( r_{\boldsymbol{\theta}} + \gamma \mathrm{E}_{\stateN' \sim p(\stateN'|\stateN, \actionN)} V^{\text{soft},\boldsymbol{\theta}}(\stateN') - V^{\text{soft},\boldsymbol{\theta}}(\stateN) \right) \right] \\
&= \mathrm{E}_{\stateN,\actionN \sim q(\stateN,\actionN)} \left[ r_{\boldsymbol{\theta}} \right] + (1-\gamma) \sum_{t=1}^{\infty} \gamma^{t} \mathrm{E}_{\stateN \sim q_{t}(\stateN)} \left[ V^{\text{soft},\boldsymbol{\theta}}(\stateN) \right] - (1-\gamma) \sum_{t=0}^{\infty} \gamma^{t} \mathrm{E}_{\stateN \sim q_{t}(\stateN)} \left[ V^{\text{soft},\boldsymbol{\theta}}(\stateN) \right] \\
&=  \mathrm{E}_{\stateN,\actionN \sim q(\stateN,\actionN)} \Big[ r_{\boldsymbol{\theta}} \Big]  - (1-\gamma) \mathrm{E}_{\stateN \sim p_{0}(\stateN)} \left[ V^{\text{soft},\boldsymbol{\theta}}(\stateN) \right].
\end{aligned}
\end{equation}
As shown by~\citet{Ziebart2010a}, the gradient of the log-likelihood (Eq.~\ref{nail_eq:maxentLL}) is given by
\begin{equation}
\begin{aligned}
\label{nail_eq:maxent_gradient}
\frac{d\mathcal{L}}{d{\boldsymbol{\theta}}} &= \mathrm{E}_{\stateN,\actionN \sim q(\stateN,\actionN)} \Big[ \frac{dr_{\boldsymbol{\theta}}}{d\boldsymbol{\theta}} \Big] - \mathrm{E}_{
	\stateN,\actionN \sim p^{\boldsymbol{\theta}}(\stateN,\actionN)
} \Big[ \frac{dr_{\boldsymbol{\theta}}}{d\boldsymbol{\theta}} \Big] \\
&= \mathrm{E}_{\stateN,\actionN \sim q(\stateN,\actionN)} \Big[ \mathbf{f}(\stateN,\actionN) \Big] - \mathrm{E}_{
	\stateN,\actionN \sim p^{\boldsymbol{\theta}}(\stateN,\actionN)
} \Big[ \mathbf{f}(\stateN, \actionN) \Big],
\end{aligned}
\end{equation}
where $p^{\boldsymbol{\theta}}(\stateN,\actionN)$ is the state-action distribution induced by the optimal policy of the entropy regularized reinforcement learning problem (Eq.~\ref{nail_eq:maxent_rl}) for the reward function $r_{\boldsymbol{\theta}}$. The first term of the gradient (Eq.~\ref{nail_eq:maxent_gradient}) can be approximated based on samples from the expert distribution. For estimating the second term one can either learn the optimal policy at each iteration~\citep{Ziebart2010a}---which is only feasible for simple problems such as low-dimensional discrete problems~\citep{Levine2011} or linear-quadratic regulators~\citep{Monfort2015}---or employ importance sampling~\citep{Boularias2011, Finn2016}. 

\subsubsection{Adversarial Inverse Reinforcement Learning}
\label{nail_sec:airl}
\gls{AIRL} is based on an adversarial formulation and models the reward function $r_{\boldsymbol{\theta}}$ as a neural network which is trained as a special type of discriminator. \citet{Fu2018} show that the gradient of their discriminator objective coincides with the maximum likelihood gradient~(Eq.~\ref{nail_eq:maxent_gradient}) when the generator is optimal, that is $p^\pi(\stateN,\actionN) = p^{\boldsymbol{\theta}}(\stateN,\actionN)$. \gls{AIRL} uses a special type of discriminator, which was suggested by~\citet{Finn2016a} for the trajectory-centric case. Namely, the logits $\nu(\stateN,\actionN)$ are given by
\begin{equation}
\label{nail_eq:airl_disc}
\nu(\stateN,\actionN) = \bar{\nu}_{\boldsymbol{\theta}}(\stateN,\actionN) - \log \pi(\actionN|\stateN)
\end{equation}
and thus directly depend on the policy $\pi$. Actually, \citet{Fu2018} propose a second modification to the discriminator so that their discriminator is given by
\begin{equation*}
\nu(\stateN,\actionN,\stateN') = f_{\boldsymbol{\theta}}(\stateN) + \gamma g(\stateN') - g(\stateN) - \log \pi(\actionN|\stateN)
\end{equation*}
which aims to recover a ``disentangled'', state-only reward function $f_{\boldsymbol{\theta}}(\stateN)$ under the assumption of deterministic system dynamics. However, the latter modification is not relevant for our theoretical analysis, and we will, thus, focus on the discriminator given by Equation~\ref{nail_eq:airl_disc}. The optimization is very similar to adversarial imitation learning by alternating between updating the discriminator using the binary cross entropy loss, and updating the policy using the discriminator logits $\nu(\stateN, \actionN)$ as reward function. As shown in Section~\ref{nail_sec:choosing_f}---and also noted by \citet{Ghasemipour2020}---such procedure would in general minimize the reverse \gls{KL} divergence. However, during the discriminator update at iteration $i$, the discriminator logits $\nu(\stateN, \actionN) = \bar{\nu}_{\boldsymbol{\theta}}(\stateN,\actionN) - \log \pi^{(i)}(\actionN|\stateN)$ are trained such that for the current policy $\pi^{(i)}$ they approximate the density ratio $\frac{q(\stateN,\actionN)}{p^{\pi^{(i)}}(\stateN,\actionN)}$. Hence, the generator update according to the adversarial formulation, would optimize the reward that is computed based on the fixed policy $\pi^{(i)}$, that is, $$r(\stateN, \actionN) = \nu(\stateN, \actionN) = \bar{\nu}_{\boldsymbol{\theta}}(\stateN,\actionN) - \log \pi^{(i)}(\actionN|\stateN).$$ Instead, \gls{AIRL} treats the discriminator as a direct function of $\pi$ by optimizing the entropy-regularized reinforcement learning problem
\begin{equation}
\label{nail_eq:airl_maxent}
J_\text{AIRL-maxent} = \underset{\pi}{\max} \int_{\stateN,\actionN} p^\pi(\stateN,\actionN) \left( \bar{\nu}_{\boldsymbol{\theta}}(\stateN,\actionN) - \log \pi(\actionN|\stateN) \right) d\stateN d\actionN,
\end{equation}
which does not correspond to the generator objective of any known adversarial formulation. \citet{Fu2018} argue that the gradient of the discriminator update corresponds to an unbiased estimate of the gradient of the \gls{MaxCausalEntIRL} objective~(Eq.~\ref{nail_eq:maxent_gradient}) when assuming that the entropy-regularized reinforcement learning problem (Eq.~\ref{nail_eq:airl_maxent}) is solved at each iteration.
However, as shown in Appendix~\ref{nail_app:bce_loss} we would not only need to assume that the policy is optimal for the given reward function $\bar{\nu}_{\boldsymbol{\theta}}$, but also that this policy matches the expert distribution $q(\stateN,\actionN)$ to show that the gradients coincide. Hence, we can only show that the optimal reward function of the maximum-likelihood objective is also a stationary point of the discriminator once the algorithm has converged---if the demonstrations can be perfectly matched---, but we can not show convergence of the algorithm by relating it to \gls{MaxCausalEntIRL}.
Hence, we argue that although \gls{AIRL} showed some promising results, its theoretical justification is currently not well understood. We will now derive a non-adversarial formulation for imitation learning (\gls{NAIL}), which is neither based on \gls{MaxCausalEntIRL} nor on generative adversarial nets, and show that adversarial inverse reinforcement learning is indeed an instance of this non-adversarial formulation.

\section{Non-Adversarial Imitation Learning}
\label{nail_sec:nail}
We aim to match the observations by minimizing the reverse \gls{KL}-divergence between the agent's and the expert's respective distribution of observations, that is,
\begin{align}
\label{nail_eq:optimizationProblem}
& \max_\pi - \int_\mathbf{o} p^\pi(\mathbf{o}) \log{\frac{p^\pi(\mathbf{o})}{q(\mathbf{o})}} d\mathbf{o} \nonumber \\
& =\max_\pi  \int_{\boldsymbol{\tau}, \mathbf{o}} p^\pi(\boldsymbol{\tau)} p(\mathbf{o}|\boldsymbol{\tau})  \log{\frac{q(\mathbf{o})}{p^\pi(\mathbf{o})}} d\boldsymbol{\tau}d\mathbf{o} \nonumber\\
& =\max_\pi  \int_{\boldsymbol{\tau}} p^\pi(\boldsymbol{\tau)} r(\boldsymbol{\tau}, \pi) d\boldsymbol{\tau}.
\end{align}
Optimization Problem~\ref{nail_eq:optimizationProblem} resembles the optimization problem solved by (episodic) reinforcement learning. However, note that the reward function 
\begin{equation}
\label{nail_eq:trajectoryReward}
r(\boldsymbol{\tau}, \pi) = \int_{\mathbf{o}} p(\mathbf{o}|\boldsymbol{\tau}) \log{\frac{q(\mathbf{o})}{\int_{\boldsymbol{\tau}} p^\pi(\boldsymbol{\tau}) p(\mathbf{o}|\boldsymbol{\tau}) d\boldsymbol{\tau}}} d\mathbf{o}   
\end{equation} depends on the agent's policy via its induced trajectory distribution $p^\pi(\boldsymbol{\tau})$. The dependency on the current policy has two main drawbacks. Firstly, the reward function can only be used to minimize the \gls{KL}-divergence if we perform sufficiently small policy updates. Secondly, the reward function given by Equation~\ref{nail_eq:trajectoryReward} does not solve the inverse reinforcement learning problem, since maximizing the expected reward will in general not match the expert distribution.
We will now show that we can remove this dependency by formulating a lower bound on Optimization Problem~\ref{nail_eq:optimizationProblem}. 
This lower bound (which corresponds to a negated upper bound on the reverse \gls{KL}) enables us to replace $p^\pi(\boldsymbol{\tau})$ in Equation~\ref{nail_eq:trajectoryReward} by a fixed, auxiliary distribution $\tilde{p}(\boldsymbol{\tau})$. In Section~\ref{nail_sec:iterativeTightening}, we will further show that this lower bound can be used to maximize the original objective given by Equation~\ref{nail_eq:optimizationProblem}.

\subsection{An Upper Bound on the Reverse \gls{KL}}
In order to derive the lower bound, we express $p^\pi(\boldsymbol{\tau})$ in terms of an auxiliary distribution $\tilde{p}(\boldsymbol{\tau})$ as follows:
\begin{align}
\label{eq:decomposition}
p^\pi(\mathbf{o}) &= \frac{p^\pi(\boldsymbol{\tau}) p(\mathbf{o}|\boldsymbol{\tau})}{p^\pi(\boldsymbol{\tau|\mathbf{o}})} \nonumber
= \frac{p^\pi(\boldsymbol{\tau}) p(\mathbf{o}|\boldsymbol{\tau})}{\tilde{p}(\boldsymbol{\tau|\mathbf{o}})} \frac{\tilde{p}(\boldsymbol{\tau|\mathbf{o}})}{p^\pi(\boldsymbol{\tau|\mathbf{o}})} = \tilde{p}(\mathbf{o}) \frac{p^\pi(\boldsymbol{\tau}) p(\mathbf{o}|\boldsymbol{\tau})}{\tilde{p}(\boldsymbol{\tau})p(\mathbf{o}|\boldsymbol{\tau})}
\frac{\tilde{p}(\boldsymbol{\tau|\mathbf{o}})}{p^\pi(\boldsymbol{\tau|\mathbf{o}})} \nonumber\\
& = \tilde{p}(\mathbf{o}) \frac{p^\pi(\boldsymbol{\tau})}{\tilde{p}(\boldsymbol{\tau})}
\frac{\tilde{p}(\boldsymbol{\tau|\mathbf{o}})}{p^\pi(\boldsymbol{\tau|\mathbf{o}})}.
\end{align}

Based on Equation~\ref{eq:decomposition} we can express Optimization Problem~\ref{nail_eq:optimizationProblem} in terms of the auxiliary distribution, that is,
\begin{align}
\label{nail_eq:reformulatedObjective}
& \max_\pi  \int_{\boldsymbol{\tau}, \mathbf{o}} p^\pi(\boldsymbol{\tau)} p(\mathbf{o}|\boldsymbol{\tau}) \log{\frac{q(\mathbf{o})}{p^\pi(\mathbf{o})}}  d\boldsymbol{\tau}d\mathbf{o} \nonumber\\
=&\max_\pi \int_{\boldsymbol{\tau}, \mathbf{o}} p^\pi(\boldsymbol{\tau)} p(\mathbf{o}|\boldsymbol{\tau}) \left( \log{\frac{q(\mathbf{o})}{\tilde{p}(\mathbf{o})}} - \log\frac{p^\pi(\boldsymbol{\tau})}{\tilde{p}(\boldsymbol{\tau})} + \log\frac{p^\pi(\boldsymbol{\tau|\mathbf{o}})}{\tilde{p}^\pi(\boldsymbol{\tau|\mathbf{o}})} \right)  d\boldsymbol{\tau}d\mathbf{o} \nonumber \\
=&\max_\pi  \int_{\boldsymbol{\tau}, \mathbf{o}} p^\pi(\boldsymbol{\tau)} p(\mathbf{o}|\boldsymbol{\tau})  \log{\frac{q(\mathbf{o})}{\tilde{p}(\mathbf{o})}} d\boldsymbol{\tau}d\mathbf{o} - \text{KL}\left(p^\pi(\boldsymbol{\tau})||\tilde{p}(\boldsymbol{\tau})\right)  \\
&  + \mathrm{E}_{p^\pi(\mathbf{o})} \left[ \text{KL}\left(p^\pi(\boldsymbol{\tau}|\mathbf{o})||\tilde{p}(\boldsymbol{\tau}|\mathbf{o})\right) \right]  \nonumber \\
=&\max_\pi \int_{\boldsymbol{\tau}, \mathbf{o}} p^\pi(\boldsymbol{\tau)} p(\mathbf{o}|\boldsymbol{\tau})  \log{\frac{q(\mathbf{o})}{\tilde{p}(\mathbf{o})}} d\boldsymbol{\tau}d\mathbf{o}  \nonumber\\
&- (1-\gamma) \sum_t \gamma^t \int_\mathbf{s} p_t^\pi(\mathbf{s}) \int_\mathbf{a} \pi(\mathbf{a}|\mathbf{s}) \log\frac{\pi(\mathbf{a}|\mathbf{s})}{\tilde{\pi}(\mathbf{a}|\mathbf{s})} d\mathbf{a}d\mathbf{s} + \mathrm{E}_{p^\pi(\mathbf{o})} \left[ \text{KL}\left(p^\pi(\boldsymbol{\tau}|\mathbf{o})||\tilde{p}(\boldsymbol{\tau}|\mathbf{o})\right) \right]  \nonumber \\
\begin{split}
\label{nail_eq:lowerbound_plus_EKL}
=&\max_\pi H(\pi) + \int_{\boldsymbol{\tau}, \mathbf{o}} p^\pi(\boldsymbol{\tau)} p(\mathbf{o}|\boldsymbol{\tau})  \log{\frac{q(\mathbf{o})}{\tilde{p}(\mathbf{o})}} d\boldsymbol{\tau}d\mathbf{o}   \\&+  (1-\gamma) \sum_t \gamma^t \int_\mathbf{s} p_t^\pi(\mathbf{s}) \int_\mathbf{a} \pi(\mathbf{a}|\mathbf{s}) \log{\tilde{\pi}(\mathbf{a}|\mathbf{s})} d\mathbf{a}d\mathbf{s}  + \mathrm{E}_{p^\pi(\mathbf{o})} \left[ \text{KL}\left(p^\pi(\boldsymbol{\tau}|\mathbf{o})||\tilde{p}(\boldsymbol{\tau}|\mathbf{o})\right) \right].
\end{split}
\end{align}
Here $H(\pi)$ denotes the discounted causal entropy of policy $\pi$, which is defined as 
\begin{equation*}
H(\mathbf{\pi}) = - (1-\gamma) \sum_{t=0}^\infty \gamma^t \int_\mathbf{s} p_t^\pi(\mathbf{s}) \int_\mathbf{a} \pi(\mathbf{a}|\mathbf{s}) \log\pi(\mathbf{a}|\mathbf{s}) d\mathbf{a} d\mathbf{s}.
\end{equation*}
As the last term of Eq.~\ref{nail_eq:lowerbound_plus_EKL} is an expected \gls{KL} divergence and thus non-negative, we can omit it to obtain the lower bound
\begin{equation}
\label{eq:lowerBoundTraj}
\mathcal{L} = H(\pi) + \int_{\mathbf{o}} p^\pi(\mathbf{o}) \log{\frac{q(\mathbf{o})}{\tilde{p}(\mathbf{o})}} d\mathbf{o} + (1-\gamma) \sum_t \gamma^t \int_\mathbf{s} p_t^\pi(\mathbf{s}) \int_\mathbf{a} \pi(\mathbf{a}|\mathbf{s}) \log{\tilde{\pi}(\mathbf{a}|\mathbf{s})} d\mathbf{a}d\mathbf{s}.
\end{equation}
For step-based, noiseless observations of the states and action, this corresponds to an entropy-regularized reinforcement learning problem
\begin{align}
\label{nail_eq:lb_objective}
\underset{\pi}{\max}\; J_{\text{NAIL}, \tilde{\pi}}(\pi) =& \underset{\pi}{\max} \int_{\stateN,\actionN} p^\pi(\stateN,\actionN) \Big( \underbrace{\log \frac{q(\stateN,\actionN)}{p^{\tilde{\pi}}(\stateN,\actionN)} + \log \tilde{\pi}(\actionN|\stateN)}_{r_\text{lb}^{\tilde{\pi}}(\stateN, \actionN)} - \log \pi(\actionN|\stateN) \Big) d\stateN d\actionN.
\end{align}

In the following we will only consider this step based setting. However, the analysis presented in this section (including Theorem~\ref{NAIL_THM:NAIL_ALG}) can be straightforwardly extended to general observations.

\begin{lemma}
	\label{NAIL_LEMMA:LOWERBOUND}
	Let $r_{\text{lb}}^{\tilde{\pi}}(\stateN, \actionN) = \log\left(\frac{q(\stateN,\actionN)}{p^{\tilde{\pi}}(\stateN,\actionN)}\right) + \log \tilde{\pi}(\actionN|\stateN)$ denote the lower bound reward function for policy $\tilde{\pi}$. Any policy $\pi$ that improves on the lower bound objective (Eq.~\ref{nail_eq:lb_objective}) compared to $\tilde{\pi}$, also decreases the reverse Kullback-Leibler divergence to the expert distribution, that is,
	\begin{equation*}
	J_{\text{NAIL}, \tilde{\pi}}(\pi) >  J_{\text{NAIL}, \tilde{\pi}}(\tilde {\pi}) \implies D_\text{RKL}(p^{\pi}(\stateN,\actionN)||q(\stateN, \actionN)) < D_\text{RKL}(p^{\tilde{\pi}}(\stateN,\actionN)||q(\stateN, \actionN)).
	\end{equation*}
	\begin{proof}
		See Appendix~\ref{nail_app:proof_nail_lb}. The proof is similar to the proof for expectation-maximization presented by \citet{Bishop2006}. Namely, since the lower bound $J_{\text{NAIL}, \tilde{\pi}}(\pi)$ is tight for the policy $\tilde{\pi}$, improving on the lower bound also increases the expected \gls{KL} term that had been omitted for deriving the lower bound.
	\end{proof}
\end{lemma}

Comparing the policy objective for the non-adversarial formulation (Eq.~\ref{nail_eq:lb_objective}) with the adversarial objective for the reverse \gls{KL} (following Eq.~\ref{nail_eq:adv_step} and Eq.~\ref{nail_eq:adversarial_reward}),
\begin{equation*}
\underset{\pi}{\max}\; J_{\text{AIL-RKL}}(\pi) = \int_{\stateN, \actionN} p^{\pi}(\stateN, \actionN) \underbrace{\log \frac{q(\stateN,\actionN)}{p^{\pi}(\stateN,\actionN)} }_{r_\text{adv}(\stateN, \actionN)} d\stateN d\actionN,
\end{equation*}
we can identify the following differences. Due to the last term in Eq.~\ref{nail_eq:lb_objective}, the reinforcement learning problem turned into an entropy regularized reinforcement learning problem. Furthermore, the lower bound reward function $r_{\text{lb}}^{\tilde{\pi}}(\stateN, \actionN)$ contains an additional term, namely, $\log{\tilde{\pi}(\actionN | \stateN)}$. Together, these changes correspond to an additional \gls{KL} objective that penalizes deviations from the policy $\tilde{\pi}$ that was used for training the discriminator. Furthermore, the density-ratio in the lower bound reward (and consequently also the lower bound reward itself) explicitly depends on the auxiliary policy $\tilde{\pi}$ and not on the policy $\pi$ that is optimized. This key difference to the adversarial formulation, enables us to greedily optimize the policy with respect to the lower bound reward function $r_{\text{lb}}^{\tilde{\pi}}(\stateN, \actionN)$ and ensures that, upon convergence, the lower bound reward function solves the inverse reinforcement learning problem.

Our upper bound on the reverse \gls{KL} divergence is similar to bounds that have been previously applied for learning hierarchical models for variational inference~\citep{Agakov2004, Tran2016, Ranganath2016, Maaloe2016, Arenz2018} and \gls{iprojection}-based density estimation~\citep{Becker2020}. Our derivations are most related to the work by~\citet{Becker2020} since we only assume samples from the target distribution $q(\obN)$. However, in contrast to~\citet{Becker2020}, we consider time-series data and do not use the bound for learning a hierarchical model, but for learning a reward function that is independent of the current policy.

\subsection{Iteratively Tightening the Bound}
\label{nail_sec:iterativeTightening}
Based on Lemma~\ref{NAIL_LEMMA:LOWERBOUND} we propose a framework for non-adversarial imitation learning that is sketched by Algorithm~\ref{nail_alg:nail}. Theorem~\ref{NAIL_THM:NAIL_ALG} shows that Algorithm~\ref{nail_alg:nail} solves the imitation learning problem when assuming that the density-ratio estimator (discriminator) is optimal. Although this assumption is very strong, it is also typical for adversarial methods~\citep{Goodfellow2014, Nowozin2016}. An important difference to the adversarial formulation stems from the fact, that we can show convergence for any policy improvement, that is, we do not require  ``sufficiently small'' generator updates. However, the algorithmic differences compared to the adversarial formulation are quite modest: the generator update of \gls{NAIL} has an additional reward term $\log \tilde{\pi}(\actionN|\stateN)$ and solves an entropy-regularized reinforcement learning problem. Together these changes correspond to an additional \gls{KL}-penalty, $\text{KL}\left(p^\pi(\boldsymbol{\tau})||\tilde{p}(\boldsymbol{\tau})\right)$ (see Eq.~\ref{nail_eq:reformulatedObjective}), that penalizes large deviations from the policy $\tilde{\pi}$ that was used for training the discriminator. Due to the similarity of both formulations, it can be difficult to show significant differences between the adversarial and the non-adversarial approach in practice. However, we will show in the next section that the non-adversarial formulation can help in better understanding existing (adversarial) methods and that it can be used to derive novel algorithms.

\begin{algorithm}[t]
	\caption{Non-Adversarial Imitation Learning}
	\label{nail_alg:nail}
	\begin{algorithmic}[1]
		\Require Expert demonstrations $\mathcal{D}=\left\{(\state{i},
		\action{i})\right\}_{i=1 \dots N}$
		\Require Initial policy $\tilde{\pi}$
		\Function{NAIL}{}
		\Repeat
		\State $\tilde{\phi}(\stateN, \actionN)=${\sc{DensityRatioEstimator}}$(\tilde{\pi}, \mathcal{D})$  \Comment{Estimate $\frac{q(\stateN,\actionN)}{p^{\tilde{\pi}}(\stateN,\actionN)}$, e.g. using samples from $p^{\tilde{\pi}}$}
		\State $\tilde{\pi} \gets $ {\sc{PolicyImprovement}}$(\tilde{\pi}, \tilde{\phi}(\stateN, \actionN))$ \Comment{improve on $J_{\text{NAIL}, \tilde{\pi}}$ using $\tilde{\phi}$ to estimate $r_{\text{lb}}^{\tilde{\pi}}$} 
		\Until{converged}
		\EndFunction
	\end{algorithmic}
\end{algorithm}

\begin{theorem}
	\label{NAIL_THM:NAIL_ALG}
	When ignoring approximation errors of the density-ratio estimator, Algorithm~\ref{nail_alg:nail} converges to a stationary point of the reverse \gls{KL} imitation learning objective.
	\begin{proof}
		See Appendix~\ref{nail_app:proof_nail_alg}
	\end{proof}
\end{theorem}

\section{Instances of Non-Adversarial Imitation Learning}
\label{nail_sec:instances}
We will discuss two instances of non-adversarial imitation learning. In Section~\ref{nail_sec:AirlAsNail}, we will focus on an existing method, \gls{AIRL}, and show that it is an instance of non-adversarial imitation learning even though it has been originally formulated as adversarial method. In Section~\ref{nail_sec:onail}, we present a novel algorithm based on our non-adversarial formulation, which is suitable for offline imitation learning.

\subsection{Adversarial Inverse Reinforcement Learning}
\label{nail_sec:AirlAsNail}
We showed in Section~\ref{nail_sec:airl} that \gls{AIRL} is neither a typical adversarial algorithm nor can it be derived from maximum causal entropy \gls{IRL}. 
However, \gls{AIRL} can be straightforwardly justified as an instance of \gls{NAIL}, as shown in Corollary~\ref{nail_corollary:AIRLisNAIL}.
\begin{corollary}
	\label{nail_corollary:AIRLisNAIL}
	Adversarial inverse reinforcement learning is an instance of non-adversarial imitation learning (Algorithm~\ref{nail_alg:nail}).
	\begin{proof}
		Let $\tilde{\pi}$ denote the current policy during the discriminator update at any given iteration. As shown in Section~\ref{nail_sec:dre}, the logits of the discriminator trained with the binary cross-entropy loss approximate the log density-ratio $\log \frac{q(\stateN,\actionN)}{p^{\tilde{\pi}}(\stateN,\actionN)}$. Hence, we have
		\begin{align*}
		\nu(\stateN, \actionN) = \bar{\nu}_{\boldsymbol{\theta}}(\stateN, \actionN) - \log \tilde{\pi}(\stateN, \actionN) \approx \log \frac{q(\stateN,\actionN)}{p^{\tilde{\pi}}(\stateN,\actionN)}
		\end{align*}
		and thus
		\begin{align*}
		\bar{\nu}_{\boldsymbol{\theta}}(\stateN, \actionN) \approx \log \frac{q(\stateN,\actionN)}{p^{\tilde{\pi}}(\stateN,\actionN)} + \log \tilde{\pi}(\stateN, \actionN) = r_{\text{lb}}^{\tilde{\pi}}(\stateN, \actionN).
		\end{align*}
		The policy objective $J_\text{AIRL-maxent}(\pi)$ (Eq.~\ref{nail_eq:airl_maxent})  solved by \gls{AIRL}, thus, corresponds to an approximation of $J_{\text{NAIL}, \tilde{\pi}}(\pi)$ (Eq.~\ref{nail_eq:lb_objective}) based on the density-ratio estimator $\widetilde{\phi}(\stateN, \actionN) = \nu(\stateN, \actionN)$. Hence, \gls{AIRL} implements Algorithm~\ref{nail_alg:nail}.
	\end{proof}
\end{corollary}
Intuitively, \gls{NAIL} solves the inverse reinforcement learning problem because any policy that maximizes the lower bound reward function $r_{\text{lb}}^{\tilde{\pi}}(\stateN, \actionN)$ matches the expert demonstration at least as good as the learned policy $\tilde{\pi}$. After convergence, when $p^{\tilde{\pi}}(\stateN,\actionN) \approx q(\stateN,\actionN)$, maximizing the lower bound reward function, thus, matches the expert demonstrations. In contrast, the adversarial reward function after convergence $r_\text{adv}(\stateN,\actionN) = \log \frac{q(\stateN, \actionN)}{p^{\tilde{\pi}}(\stateN,\actionN)} \approx 0$ is specific to the current policy and converges to a constant, which is not suitable for matching the expert demonstrations.
However, as noted by \citet{Fu2018}, when modeling the expert according to Equation~\ref{nail_eq:maxent_expert_model}, the reward function learned by \gls{AIRL}/\gls{NAIL} corresponds to the advantage function $A^{\text{soft}}_{\text{expert}}(\stateN,\actionN) = \log \pi_{\text{expert}}(\actionN | \stateN)$ if the expert's distribution can be matched exactly by the agent. To be more specific, we can see from the definition of $r_{\text{lb}}^{\tilde{\pi}}(\stateN, \actionN)$ that, in general, it additionally contains a correction-term based on the mismatch of the induced state distributions, namely,
\begin{equation*}
r_{\text{lb}}^{\tilde{\pi}}(\stateN, \actionN) = \log\left(\frac{q(\stateN,\actionN)}{p^{\tilde{\pi}}(\stateN,\actionN)}\right) + \log \tilde{\pi}(\actionN|\stateN) = \log\left(\frac{q(\stateN)}{p^{\tilde{\pi}}(\stateN)}\right) + \log \pi_{\text{expert}}(\actionN| \stateN).
\end{equation*}
As the advantage function strongly depends on the system dynamics, the learned reward function is typically not transferable to different dynamical systems. Hence, \citet{Fu2018} proposed to parameterize the discriminator as
\begin{equation*}
\nu(\stateN,\actionN,\stateN') = f_{\boldsymbol{\theta}}(\stateN) + \gamma g_{\eta}(\stateN') - g_{\eta}(\stateN) - \log \pi(\actionN|\stateN)
\end{equation*}
and showed that $f_{\boldsymbol{\theta}}(\stateN)$ approximates the reward function of the expert, when assuming that the system dynamics are deterministic and that the true reward function does not depend on the actions. Note that this modification is also covered by our derivation which does not make specific assumptions on the parameterization of $\bar{\nu}(\stateN, \actionN)$.

\subsection{Offline Non-Adversarial Imitation Learning}
\label{nail_sec:onail}
We will now use the non-adversarial formulation to derive a novel algorithm for offline imitation learning. Offline imitation learning considers the imitation learning problem, with the additional restriction that we are not able to interact with the environment during learning. Behavioral cloning (BC) is a common approach for offline imitation learning that frames imitation learning as supervised learning, for example, by maximizing the likelihood of the states and actions, that is,
\begin{equation*}
\underset{\pi}{\max}\; \Big[ J_\text{bc}(\pi) = \mathrm{E}_{\stateN, \actionN \sim q(\stateN, \actionN)} \left[ \log \pi(\actionN|\stateN) \right] \Big].
\end{equation*}
However, as observed by~\citet{Pomerleau1989} and further examined by~\citet{Ross2010} compounding errors can lead to covariate shift, that is, the learned policy may reach states that significantly differ from the states encountered by the expert. As the policy was not trained on these states, it is often not able to recover from such mistakes.

\subsubsection{Interlude: ValueDice}
Our method is inspired by ValueDice~\citep{Kostrikov2020} which is able to outperform behavioral cloning for offline imitation learning. ValueDice aims to match the expert distribution $q(\stateN, \actionN)$ by expressing the reverse \gls{KL} divergence using the Donsker-Varadhan representation~\citep{Donsker1983}, that is,
\begin{equation}
\label{nail_eq:DV_IL}
\underset{\pi}{\min}\;\text{KL}(p^{\pi}(\stateN,\actionN)||q(\stateN, \actionN)) = \underset{\pi}{\min} \; \underset{\nu}{\max} \; -\log \mathrm{E}_{\stateN,\actionN \sim q(\stateN,\actionN)} \left[ \exp \left( \nu(\stateN, \actionN) \right) \right] +  \mathrm{E}_{\stateN,\actionN \sim p^{\pi}(\stateN,\actionN)} \left[ \nu(\stateN, \actionN) \right].
\end{equation}
Similar to the \gls{KLIEP} loss or the binary cross-entropy loss, the optimal solution of the inner maximization problem is given by
\begin{equation}
\label{nail_eq:optimalNuDV}
\nu^{\star}(\stateN, \actionN) = \log \frac{p^{\pi}(\stateN,\actionN)}{q(\stateN,\actionN)} + \textrm{const},
\end{equation}
which closely relates the saddle point problem given by Equation~\ref{nail_eq:DV_IL} to the previously discussed adversarial methods for minimizing the reverse \gls{KL} divergence. Indeed, a similar algorithm to ValueDice could be straightforwardly derived from the $f$-\gls{GAN} formulation for the reverse \gls{KL} (Eq.~\ref{nail_eq:fgan_rkl_objective}), which differs from Optimization Problem~\ref{nail_eq:DV_IL} by not taking the logarithm of the first expectation.

Optimization Problem~\ref{nail_eq:DV_IL} can not directly be used for offline imitation learning since we can not get samples from the agent's state-action distribution $p^{\pi}(\stateN,\actionN)$ to approximate the second expectation. Hence, \citet{Kostrikov2020} applied a trick for offline density-ratio estimation~\citep{Nachum2019} by using a change of variables to express $\nu(\stateN,\actionN)$ in terms of the Q-Function $Q^{\pi}_{\nu}$ for policy $\pi$ where the reward function is given by $\nu(\stateN,\actionN)$, that is
\begin{equation}
\label{nail_eq:VD_CoV}
\nu(\stateN,\actionN) = Q^{\pi}_{\nu}(\stateN, \actionN) - \gamma \mathrm{E}_{\stateN' \sim p(\stateN'|\stateN,\actionN)} \mathrm{E}_{\actionN' \sim \pi(\actionN'|\stateN')} \left[ Q^{\pi}_{\nu}(\stateN', \actionN')  \right].
\end{equation}
When applying the change of variables (Eq.~\ref{nail_eq:VD_CoV}) and optimizing Eq.~\ref{nail_eq:DV_IL} with respect to the Q-function, the expected ``reward'' given by the second expectation can be computed based on the initial state distribution and the Q-function, that is,
\begin{align}
\begin{split}
\nonumber
\underset{\pi}{\min} \; \underset{ Q^{\pi}_{\nu}}{\max} \; -&\log \mathrm{E}_{\stateN,\actionN \sim q(\stateN,\actionN)} \left[ \exp \left( Q^{\pi}_{\nu}(\stateN, \actionN) - \gamma \mathrm{E}_{\stateN' \sim p(\stateN'|\stateN,\actionN)} \mathrm{E}_{\actionN' \sim \pi(\actionN'|\stateN')} \left[ Q^{\pi}_{\nu}(\stateN', \actionN')  \right] \right) \right] \\
&+  \mathrm{E}_{\stateN,\actionN \sim p^{\pi}(\stateN,\actionN)} \left[ Q^{\pi}_{\nu}(\stateN, \actionN) - \gamma \mathrm{E}_{\stateN' \sim p(\stateN'|\stateN,\actionN)} \mathrm{E}_{\actionN' \sim \pi(\actionN'|\stateN')} \left[ Q^{\pi}_{\nu}(\stateN', \actionN')  \right] \right] 
\end{split}
\\
\begin{split}
\label{nail_eq:VD_obj}
=\underset{\pi}{\min} \; \underset{ Q^{\pi}_{\nu}}{\max} &-\log \mathrm{E}_{\stateN,\actionN \sim q(\stateN,\actionN)} \left[ \exp \left( Q^{\pi}_{\nu}(\stateN, \actionN) - \gamma \mathrm{E}_{\stateN' \sim p(\stateN'|\stateN,\actionN)} \mathrm{E}_{\actionN' \sim \pi(\actionN'|\stateN')} \left[ Q^{\pi}_{\nu}(\stateN', \actionN')  \right] \right) \right] \\
&+ (1-\gamma) \mathrm{E}_{\state{0} \sim p_{0}(\stateN)} \mathrm{E}_{\action{0} \sim \pi(\action{0}|\state{0})} \left[ Q^{\pi}_{\nu}(\state{0}, \action{0})  \right].
\end{split}
\end{align}
As shown by \citet{Kostrikov2020}, a (biased) Monte-Carlo estimate of Eq.~\ref{nail_eq:VD_obj} can be optimized based on triplets $(\stateN, \actionN, \stateN')$ sampled from the expert demonstrations, initial states $\state{0}$ sampled from the initial state distribution and actions $\actionN'$ and $\action{0}$ sampled from the policy. Compared to the other adversarial methods discussed earlier, ValueDice differs mainly by directly learning the Q-function for the adversarial reward $r_\text{adv}(\stateN, \actionN)=\log\left( \frac{q(\stateN, \actionN)}{p^{\pi}(\stateN, \actionN)} \right)$ which does not only make it applicable to the offline setting, but---conveniently---also greatly simplifies the reinforcement learning problem. However, common to prior methods for adversarial imitation learning, ValueDice involves solving a saddle point problem (Eq.~\ref{nail_eq:VD_obj}) which may require careful tuning of the step size for the policy updates to avoid too greedy updates.

\subsubsection{ONAIL}
Instead of framing distribution matching as a saddle-point problem, we use the non\hyp{}adversarial formulation to derive a (soft) actor critic algorithm. That is, we alternate between a critic update, where we learn the soft Q-function $Q_{\text{lb}}^{\text{soft, }\pi^{(i)}}$ for the current policy $\pi^{(i)}$ and lower bound reward function $r_{\text{lb}}^{\pi^{(i)}}(\stateN, \actionN)$, and an actor update where we use the soft Q-function to find a policy $\pi^{(i+1)}$ that improves on the lower bound objective $J_{\text{NAIL},\pi^{(i)}}$ (Eq.~\ref{nail_eq:lb_objective}). 

The soft Q-function for the lower bound reward function $r_{\text{lb}}^{\pi^{(i)}}(\stateN, \actionN)$ can be learned analogously to the (standard) Q-function for the adversarial reward function by applying a change of variables to the \gls{KLIEP}-loss (Eq.~\ref{nail_eq:fgan_rkl_objective}) or Donsker-Varadhan-loss (Eq.~\ref{nail_eq:DV_IL}). However, as shown by Lemma~\ref{NAIL_LEMMA:Q_RELATIONS}, the soft Q-function for the lower bound reward can also be directly computed from the Q-function of the adversarial reward.
\begin{lemma}
	\label{NAIL_LEMMA:Q_RELATIONS}
	Let $Q_{r}^{\tilde{\pi}}$ be the Q-function for policy $\tilde{\pi}$ and a given reward function $r$. Further, let $Q_{r_\text{lb}}^{\text{soft},\tilde{\pi}}$ 
	be the soft Q-function for policy $\tilde{\pi}$ and reward function $r_{\text{lb}}(\stateN, \actionN) = r(\stateN,\actionN) + \log \tilde{\pi}(\actionN|\stateN)$. Then, $Q_{r_\text{lb}}^{\text{soft},\tilde{\pi}}$ can be expressed in terms of $Q_{r}^{\tilde{\pi}}$ as follows:
	\begin{equation*}
	Q_{r_\text{lb}}^{\text{soft},\tilde{\pi}}(\stateN,\actionN) = Q_{r}^{\tilde{\pi}}(\stateN,\actionN) + \log \tilde{\pi}(\actionN|\stateN).
	\end{equation*}
	\begin{proof}
		see Appendix~\ref{nail_app:proof_Q_relations}.
	\end{proof}
\end{lemma}

Lemma~\ref{NAIL_LEMMA:Q_RELATIONS} might seem surprising at first because in general it is neither straightforward to transform a Q-function to a soft Q-function nor to transform the Q function for one reward function to the Q-function for a different reward function. However, intuitively, the expected negative cross-entropy to come (originating from the change of reward function) and the expected entropy to come (originating from the change to the soft Q-function) cancel out, since both terms are computed for policy $\tilde{\pi}$.
Hence, both Q functions only differ by the change of the immediate reward. Lemma~\ref{NAIL_LEMMA:Q_RELATIONS}, thus, enables us to implement the critic update by performing the inner maximization of the saddle point solved by ValueDice~\ref{nail_eq:VD_obj}, that is,
\begin{align}
\label{nail_eq:nail_critic}
\begin{split}
-Q_{r_{\text{adv}}}^{\tilde{\pi}} = \underset{Q^{\tilde{\pi}}_{\nu}}{\argmax} &-\log \mathrm{E}_{\stateN,\actionN \sim q(\stateN,\actionN)} \left[ \exp \left( Q^{\tilde{\pi}}_{\nu}(\stateN, \actionN) - \gamma \mathrm{E}_{\stateN' \sim p(\stateN'|\stateN,\actionN)} \mathrm{E}_{\actionN' \sim \tilde{\pi}(\actionN'|\stateN')} \left[ Q^{\tilde{\pi}}_{\nu}(\stateN', \actionN')  \right] \right) \right] \\
&+ (1-\gamma) \mathrm{E}_{\state{0} \sim p_{0}(\stateN)} \mathrm{E}_{\action{0} \sim \tilde{\pi}(\action{0}|\state{0})} \left[ Q^{\tilde{\pi}}_{\nu}(\state{0}, \action{0})  \right].
\end{split}
\end{align}
Please note, that we changed the sign for $Q_{r_{\text{adv}}}^{\tilde{\pi}}$ because $\nu$ (Eq.~\ref{nail_eq:optimalNuDV}) approximates the adversarial cost function $\nu(\stateN,\actionN) \approx -r_\text{adv}(\stateN,\actionN) + \mathrm{const}$.

In contrast to ValueDice, we do not need to solve a saddle point problem, since we can replace the policy update of ValueDice (based on Eq.~\ref{nail_eq:VD_obj}) with an update based on the loss
\begin{align}
\label{nail_eq:nail_actor_loss}
\mathcal{L}_{\text{actor}, \pi^{(i)}}(\pi) =& - \mathrm{E}_{\stateN \sim z(\stateN)} \left[ \int_{\actionN} \pi(\actionN|\stateN) \left( Q_{\text{adv}}^{\pi^{(i)}} + \log {\pi}^{(i)}(\actionN|\stateN) - \log \pi(\actionN|\stateN) \right) d\actionN \right] \\
=& - \mathrm{E}_{\stateN \sim z(\stateN)} \left[ \mathcal{L}_{\text{actor}, \pi^{(i)}}(\pi, \stateN) \right], \nonumber
\end{align}
which can be optimized using the reparameterization trick~\citep{Kingma2014, Rezende2014}. The state distribution $z(\stateN)$ used for training the policy should ideally have full support on every state that is encountered by the new policy $\pi$. In our experiments we found it sufficient to only use the states encountered during the expert demonstrations, that is $z(\stateN) = q(\stateN)$. However, it would also be possible to use artificial states, e.g., by adding noise to the expert demonstrations, or to build a replay buffer by rolling out the policy after each iteration (which would leave the offline-regime).
The conceptual difference between \gls{ONAIL} and ValueDice enables us to show convergence even for large improvement on the loss (Eq.~\ref{nail_eq:nail_actor_loss}) based on Lemma~\ref{NAIL_LEMMA:NAIL_ACTOR_STEP}.

\begin{lemma}
	\label{NAIL_LEMMA:NAIL_ACTOR_STEP}
	When ignoring approximation errors of the Q-function $Q^{\tilde{\pi}}_{r_{\text{adv}}}$, any policy $\pi$ that achieves lower or equal loss $\mathcal{L}_{\text{actor}, \pi^{(i)}}(\pi,\stateN)$ (Eq.~\ref{nail_eq:nail_actor_loss}) than $\pi^{(i)}$ on any state $\stateN$ also improves on $\pi^{(i)}$ with respect to the lower bound objective $J_{\text{NAIL},\pi^{(i)}}(\pi)$ (Eq.~\ref{nail_eq:lb_objective}), that is,
	\begin{equation*}
	\forall_{\stateN}:\mathcal{L}_{\text{actor}, \pi^{(i)}}(\pi, \stateN) \le \mathcal{L}_{\text{actor}, \pi^{(i)}}(\pi^{(i)}, \stateN) \implies J_{\text{NAIL},\pi^{(i)}}(\pi) \ge J_{\text{NAIL},\pi^{(i)}}(\pi^{(i)}).
	\end{equation*}
	\begin{proof}
		See Appendix~\ref{nail_app:proof_actor_step}.
	\end{proof}
\end{lemma}

The resulting algorithm (Alg.~\ref{nail_alg:onail}) is, thus, an instance of non-adversarial imitation learning based on the implicit density-ratio estimator
\begin{equation*}
\widetilde{\phi}(\stateN,\actionN) = \exp \left( Q_{r_{\text{adv}}}^{\tilde{\pi}}(\stateN, \actionN) - \gamma \mathrm{E}_{\stateN' \sim p(\stateN'|\stateN,\actionN)} \mathrm{E}_{\actionN' \sim \pi(\actionN'|\stateN')} \left[ Q_{r_{\text{adv}}}^{\tilde{\pi}}(\stateN', \actionN')  \right] \right).
\end{equation*}

\begin{theorem}
	When ignoring approximation errors of the Q-function $Q^{\tilde{\pi}}_{r_{\text{adv}}}$, Algorithm~\ref{nail_alg:onail} converges to a stationary point of the reverse KL imitation learning objective.
	\begin{proof}
		Based on Lemma~\ref{NAIL_LEMMA:NAIL_ACTOR_STEP}, the theorem can be derived analogously to Theorem~\ref{NAIL_THM:NAIL_ALG}.
	\end{proof} 
\end{theorem}
\begin{algorithm}[t]
	\caption{Offline Non-Adversarial Imitation Learning}
	\label{nail_alg:onail}
	\begin{algorithmic}[1]
		\Require Expert demonstrations $\mathcal{D}=\left\{(\state{i},
		\action{i})\right\}_{i=1 \dots N}$
		\Require Initial policy $\tilde{\pi}$
		\Function{ONAIL}{}
		\Repeat
		\State $Q_{r_{\text{adv}}}^{\tilde{\pi}} \gets $ {\sc{ImplicitDensityRatioEstimator}}$(\tilde{\pi},\mathcal{D})$ \Comment{approximately solve Optimization Problem~\ref{nail_eq:nail_critic}}
		\State $\tilde{\pi} \gets $ {\sc{PolicyImprovement}}$(\tilde{\pi}, Q_{r_{\text{adv}}}^{\tilde{\pi}})$ \Comment{improve on $J_{\text{NAIL}, \tilde{\pi}}$ by decreasing loss $\mathcal{L}_{\text{actor}, \tilde{\pi}}(\pi)$
			(Eq.~\ref{nail_eq:nail_actor_loss}) } 
		\Until{converged}
		\EndFunction
	\end{algorithmic}
\end{algorithm}

\section{Experiments}
\label{nail_sec:experiments}
\begin{figure}
	\centering
	\begin{subfigure}{.24\textwidth}
		\centering
		\includegraphics[height=76pt]{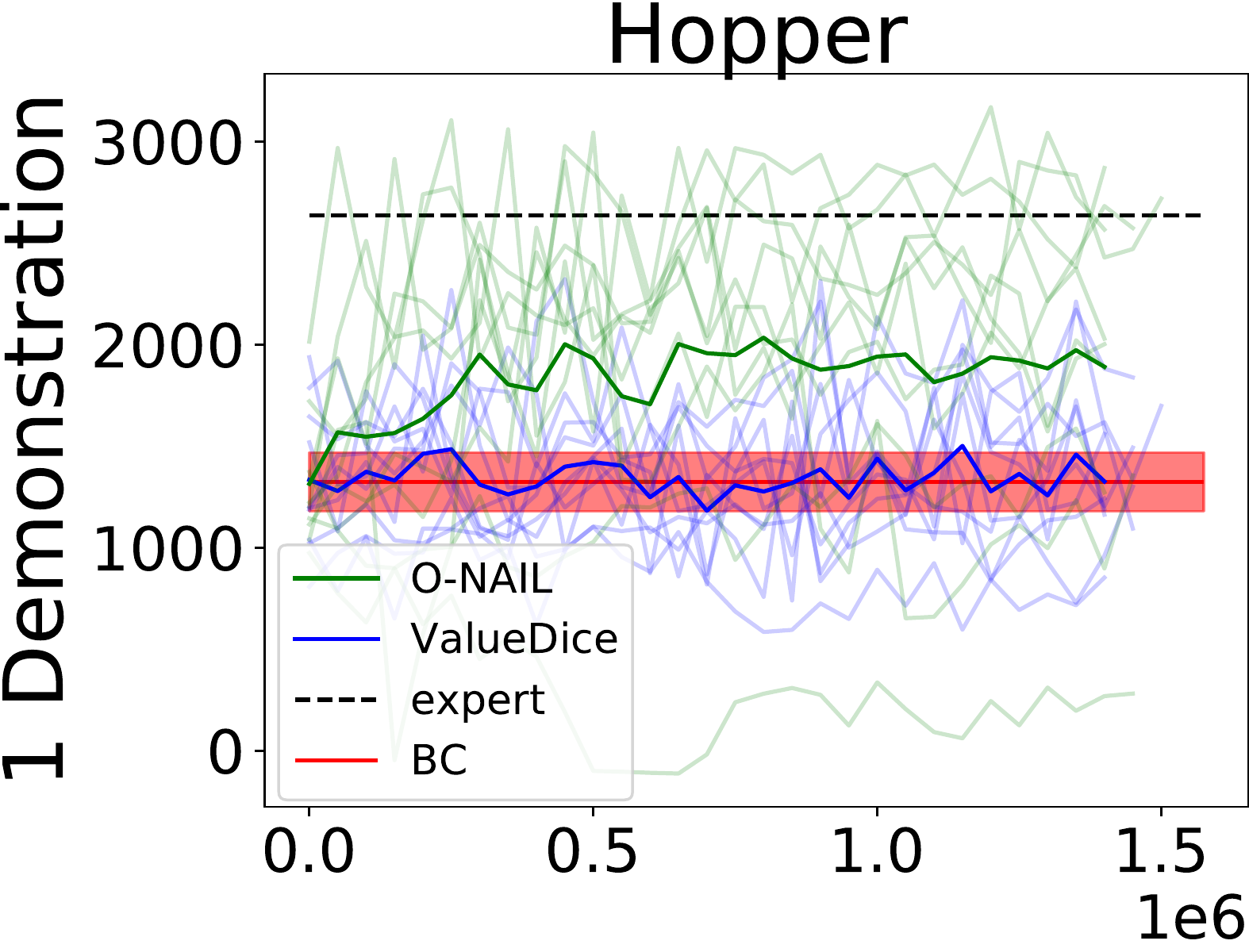}
	\end{subfigure}
	\begin{subfigure}{.24\textwidth}
		\centering
		\includegraphics[height=76pt]{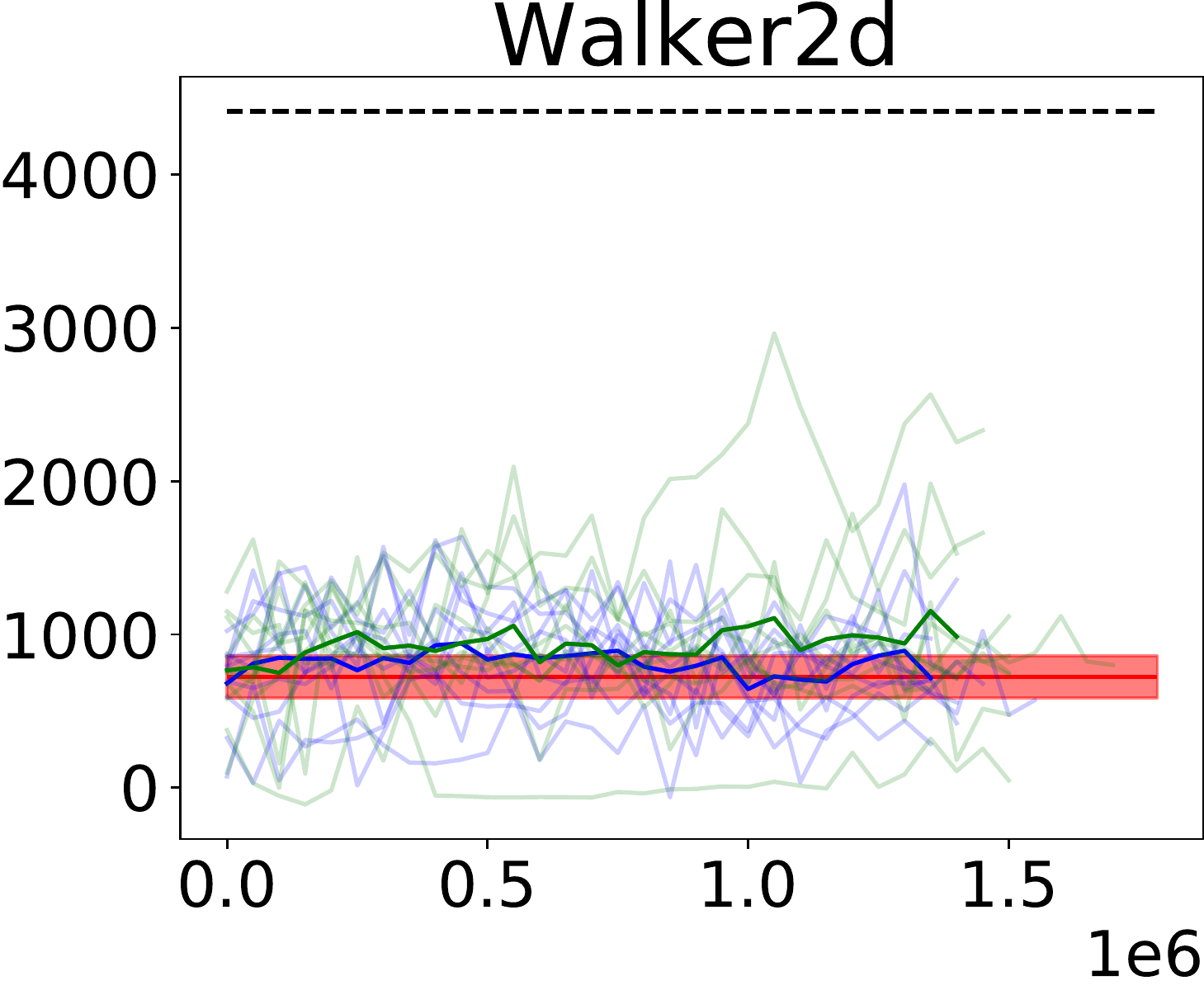}
	\end{subfigure}
	\begin{subfigure}{.24\textwidth}
		\centering
		\includegraphics[height=76pt]{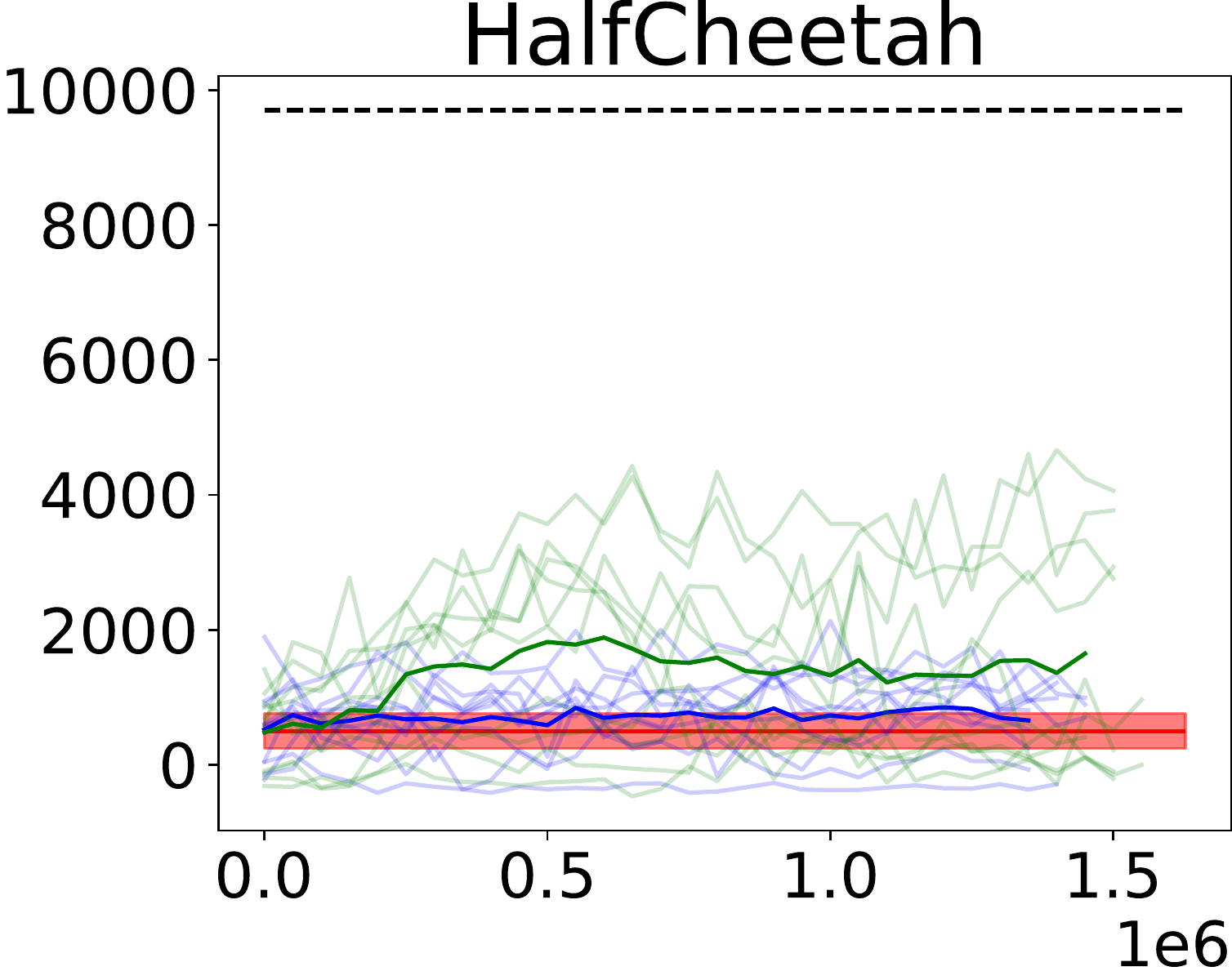}
	\end{subfigure}
	\begin{subfigure}{.24\textwidth}
		\centering
		\includegraphics[height=76pt]{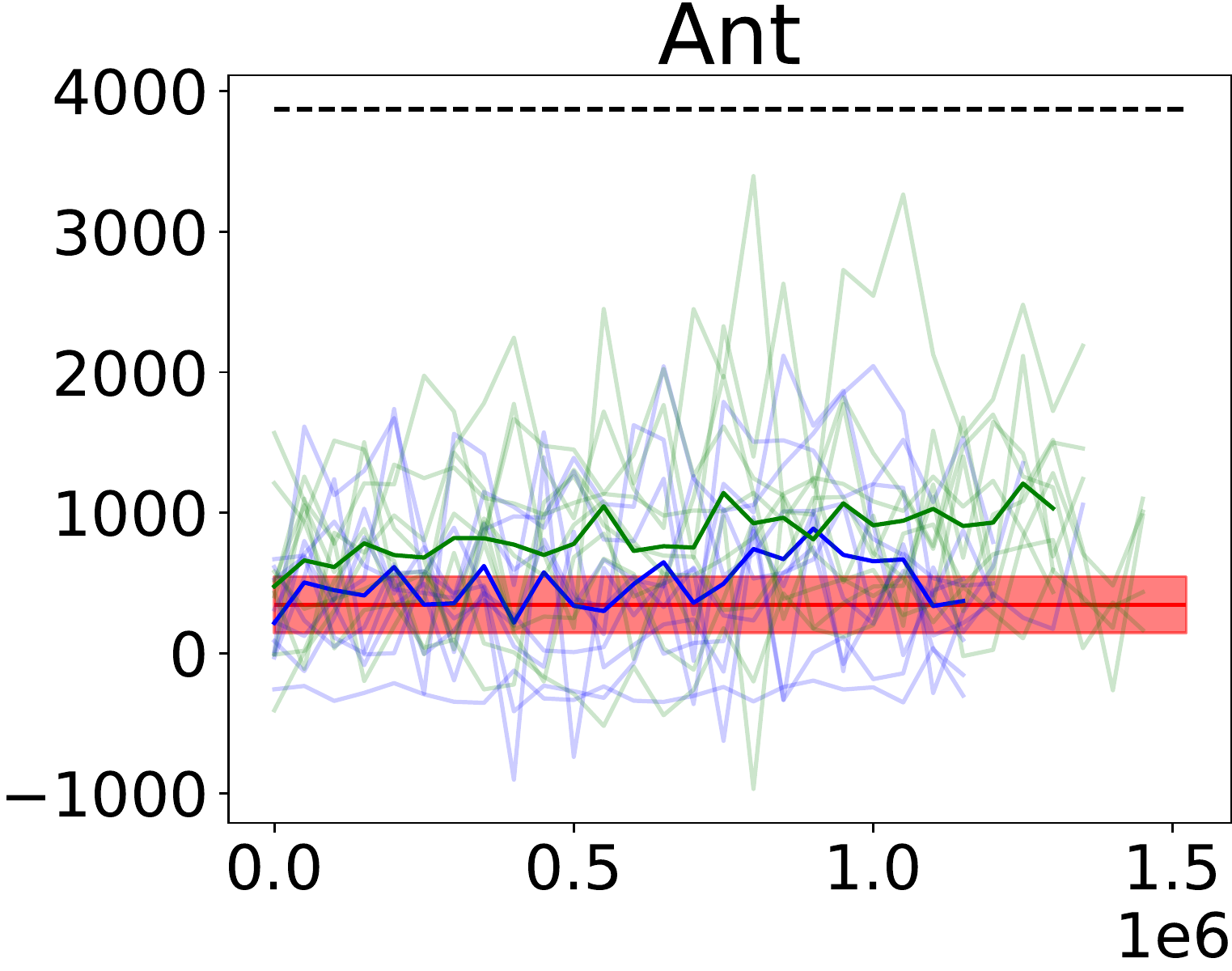}
	\end{subfigure}
	\begin{subfigure}{.24\textwidth}
		\centering
		\includegraphics[height=70pt]{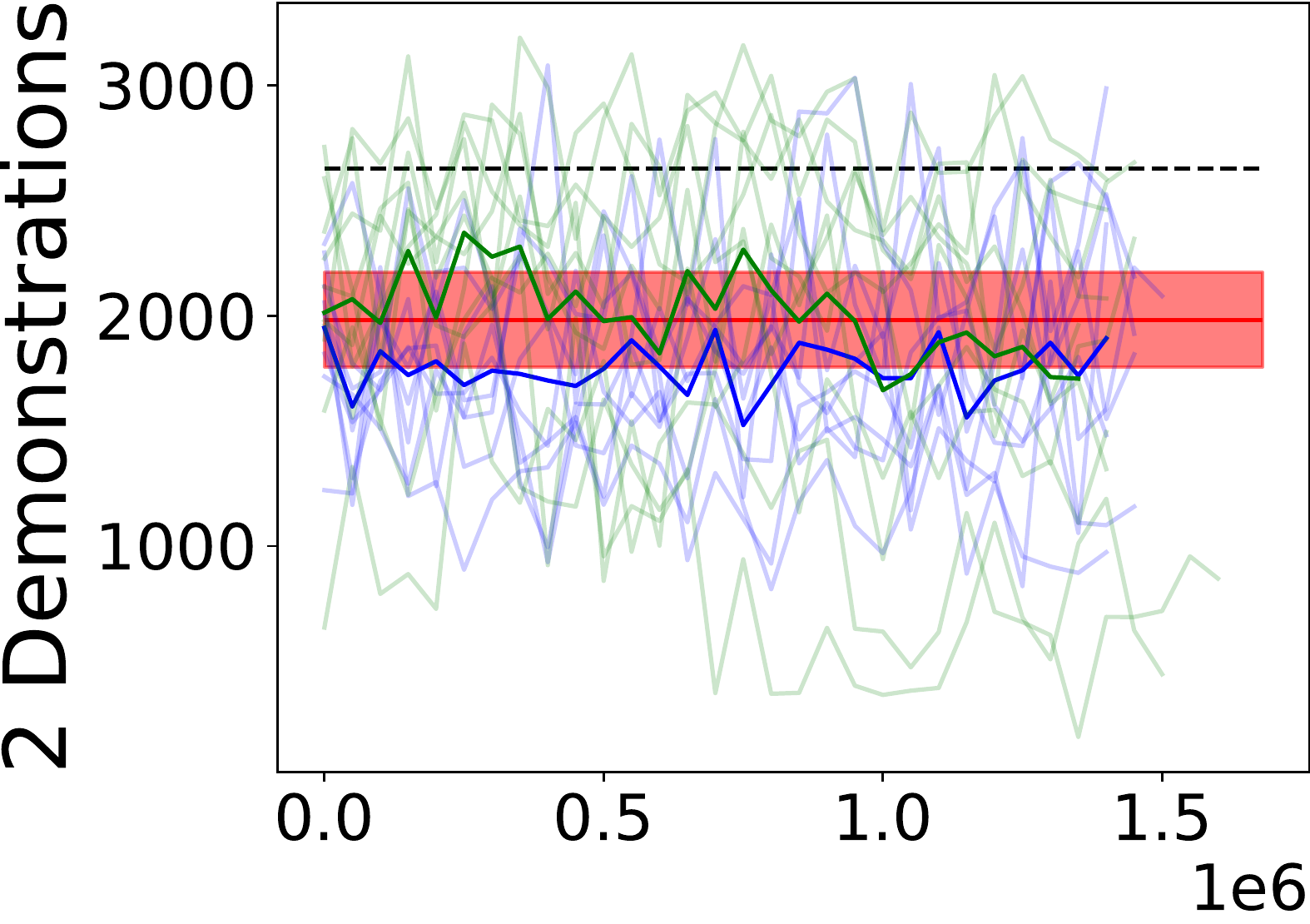}
	\end{subfigure}
	\begin{subfigure}{.24\textwidth}
		\centering
		\includegraphics[height=70pt]{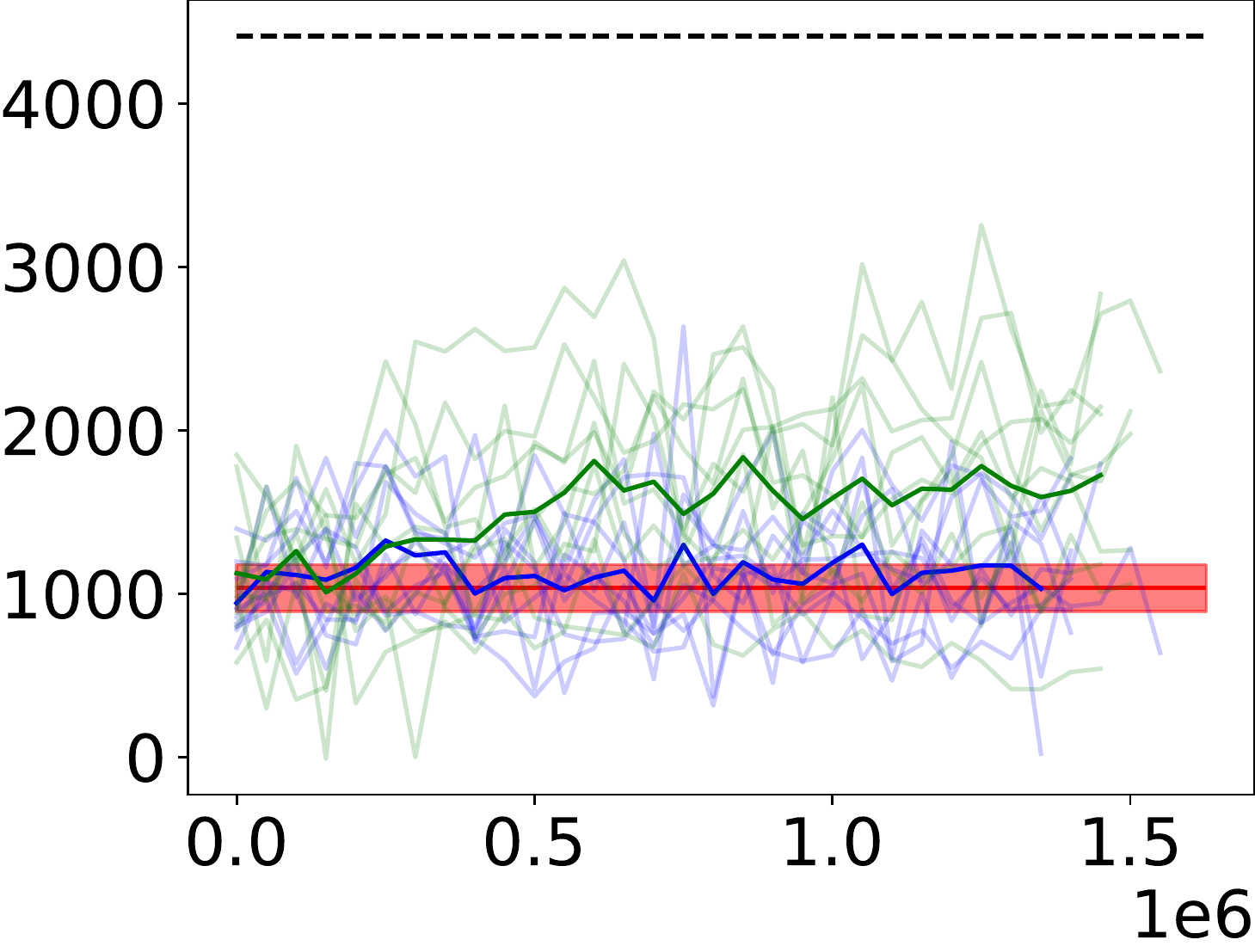}
	\end{subfigure}
	\begin{subfigure}{.24\textwidth}
		\centering
		\includegraphics[height=70pt]{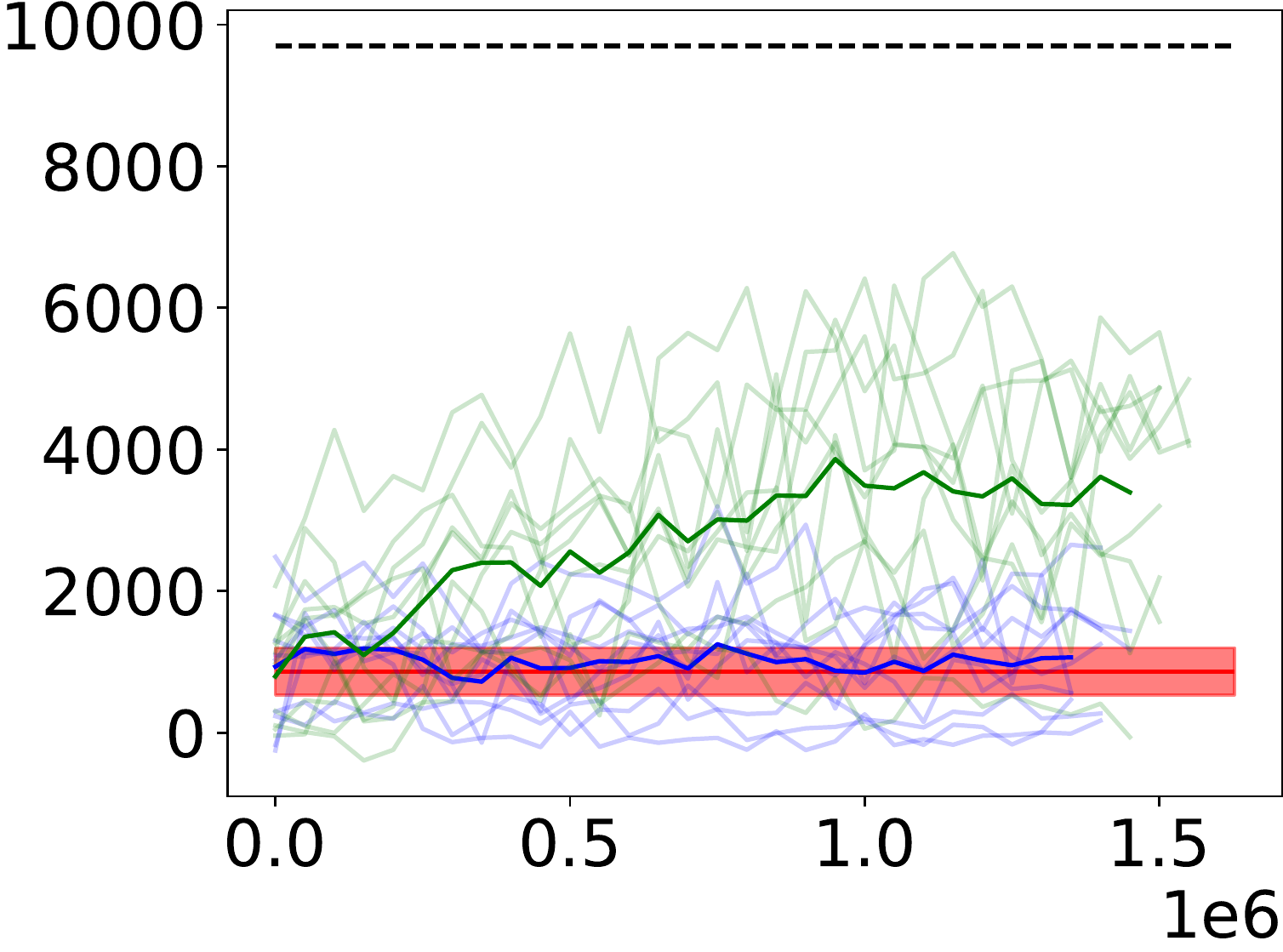}
	\end{subfigure}
	\begin{subfigure}{.24\textwidth}
		\centering
		\includegraphics[height=70pt]{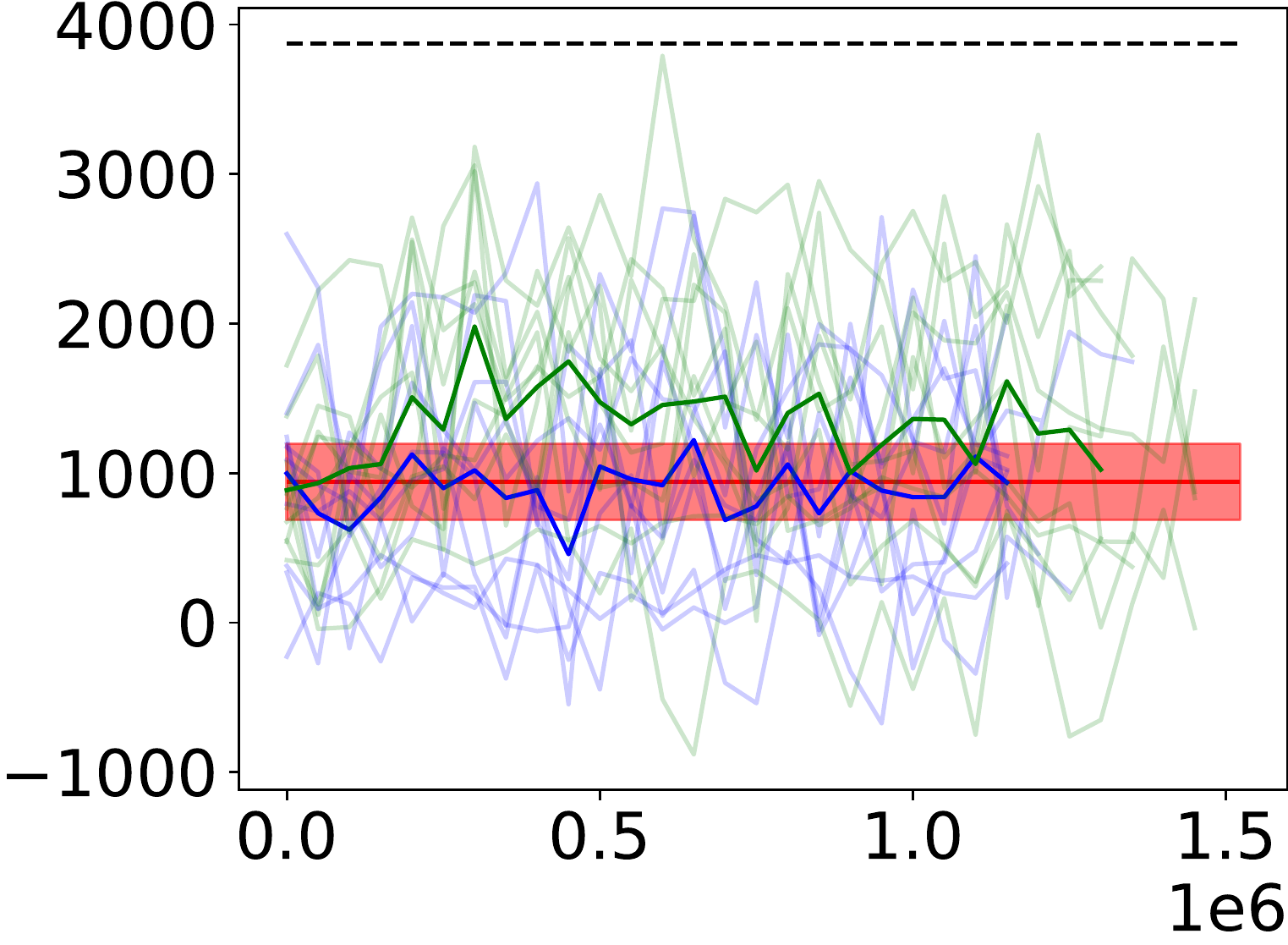}
	\end{subfigure}
	\begin{subfigure}{.24\textwidth}
		\centering
		\includegraphics[height=70pt]{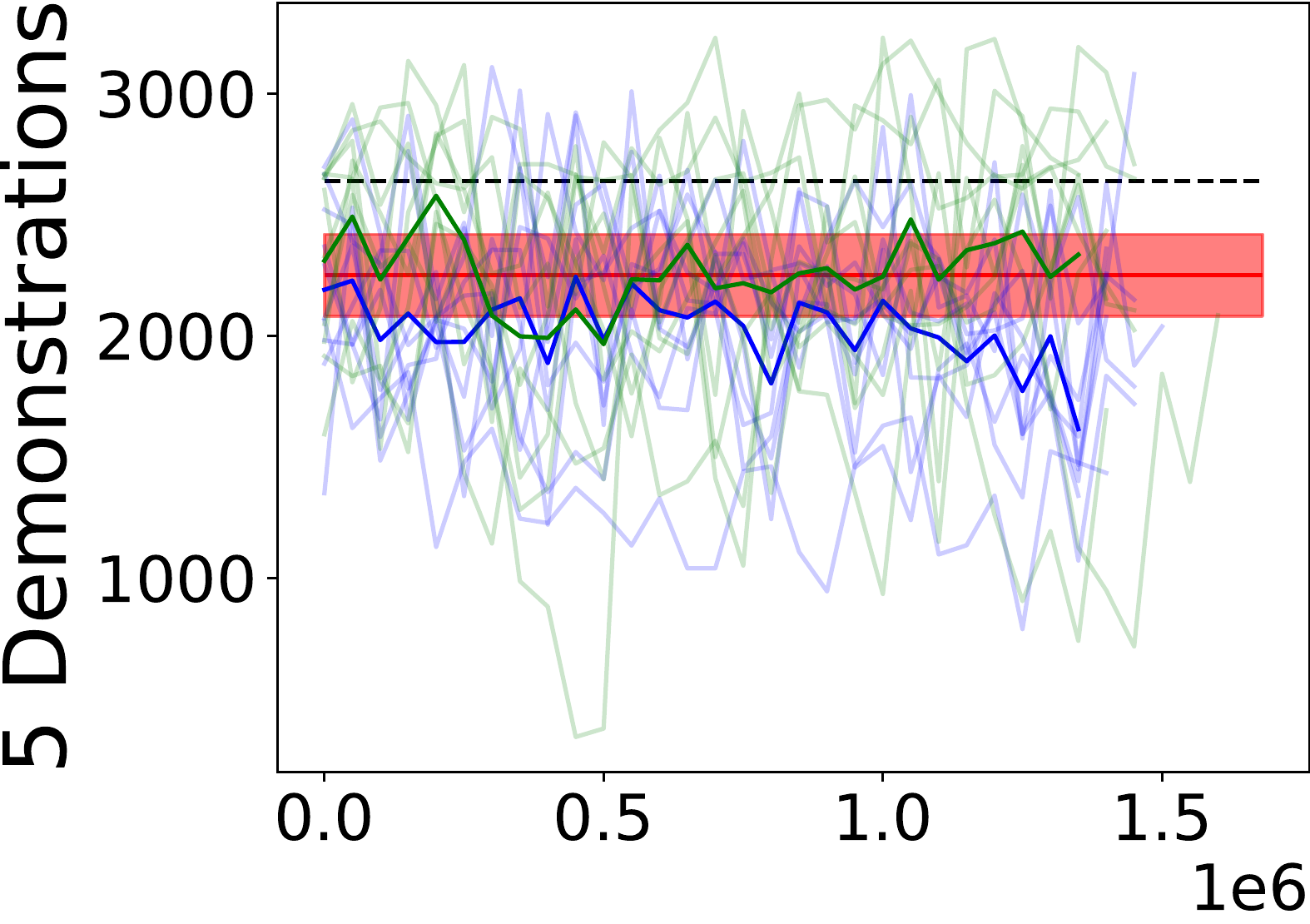}
	\end{subfigure}
	\begin{subfigure}{.24\textwidth}
		\centering
		\includegraphics[height=70pt]{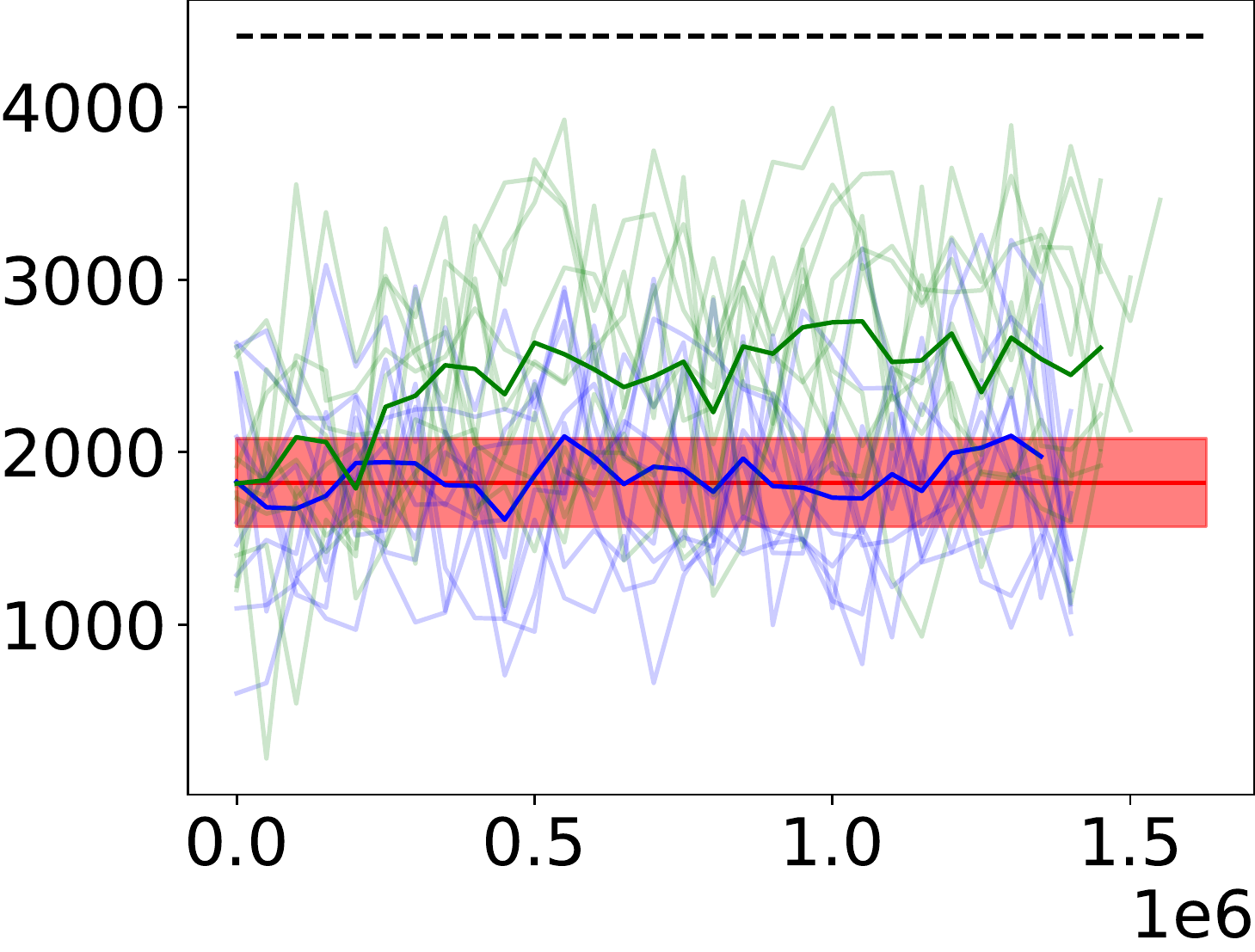}
	\end{subfigure}
	\begin{subfigure}{.24\textwidth}
		\centering
		\includegraphics[height=70pt]{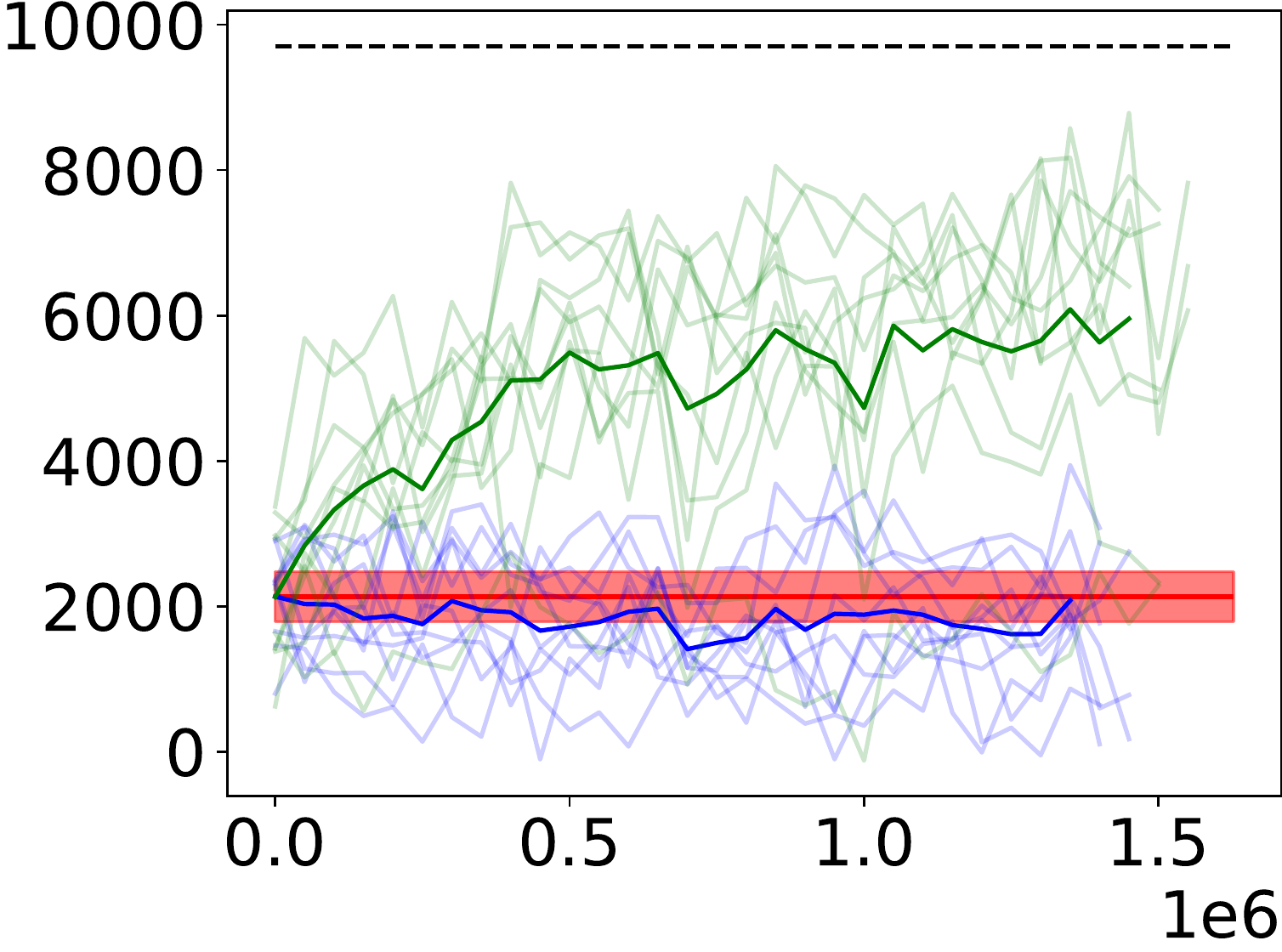}
	\end{subfigure}
	\begin{subfigure}{.24\textwidth}
		\centering
		\includegraphics[height=70pt]{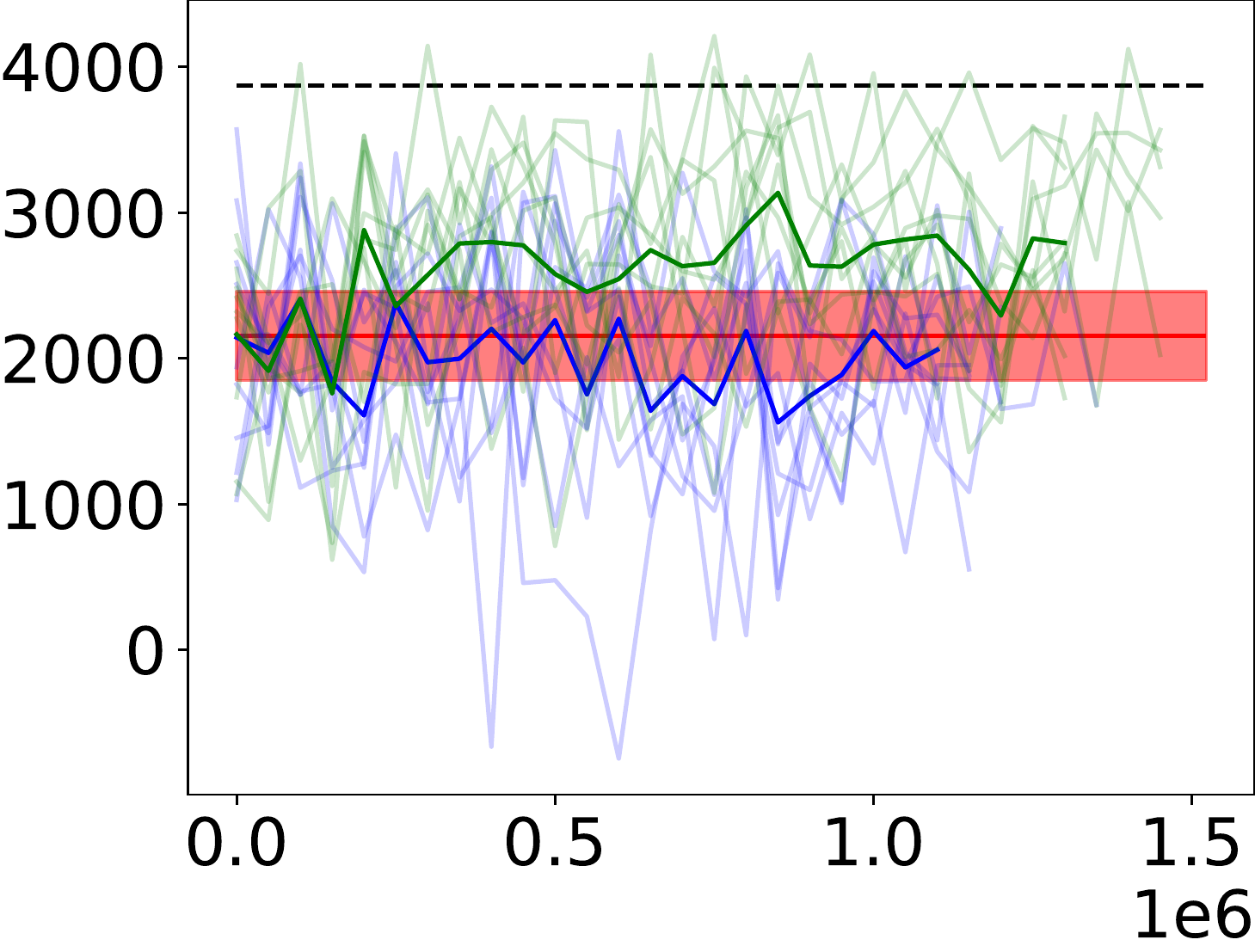}
	\end{subfigure}
	\begin{subfigure}{.24\textwidth}
		\centering
		\includegraphics[height=70pt]{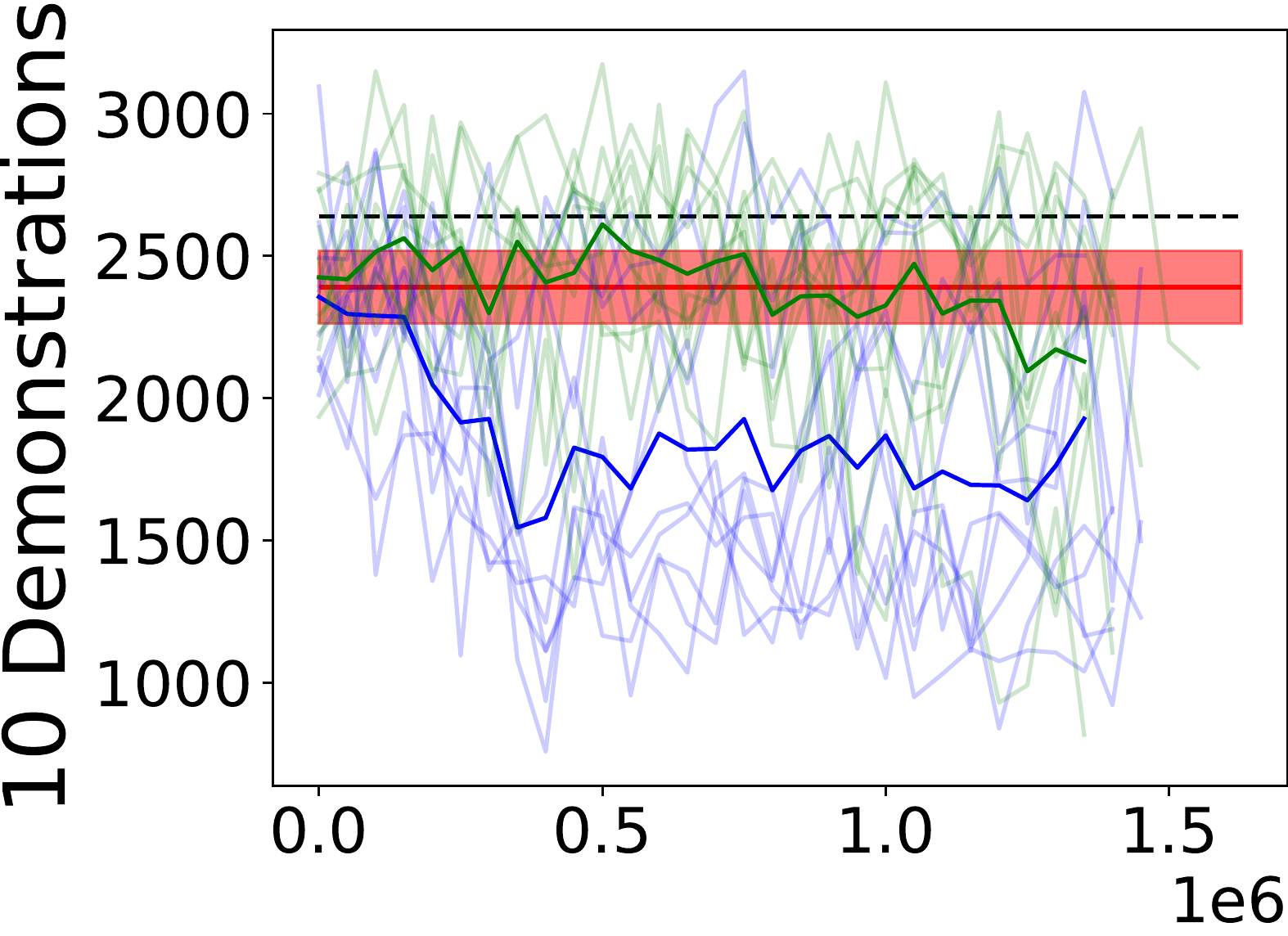}
	\end{subfigure}
	\begin{subfigure}{.24\textwidth}
		\centering
		\includegraphics[height=70pt]{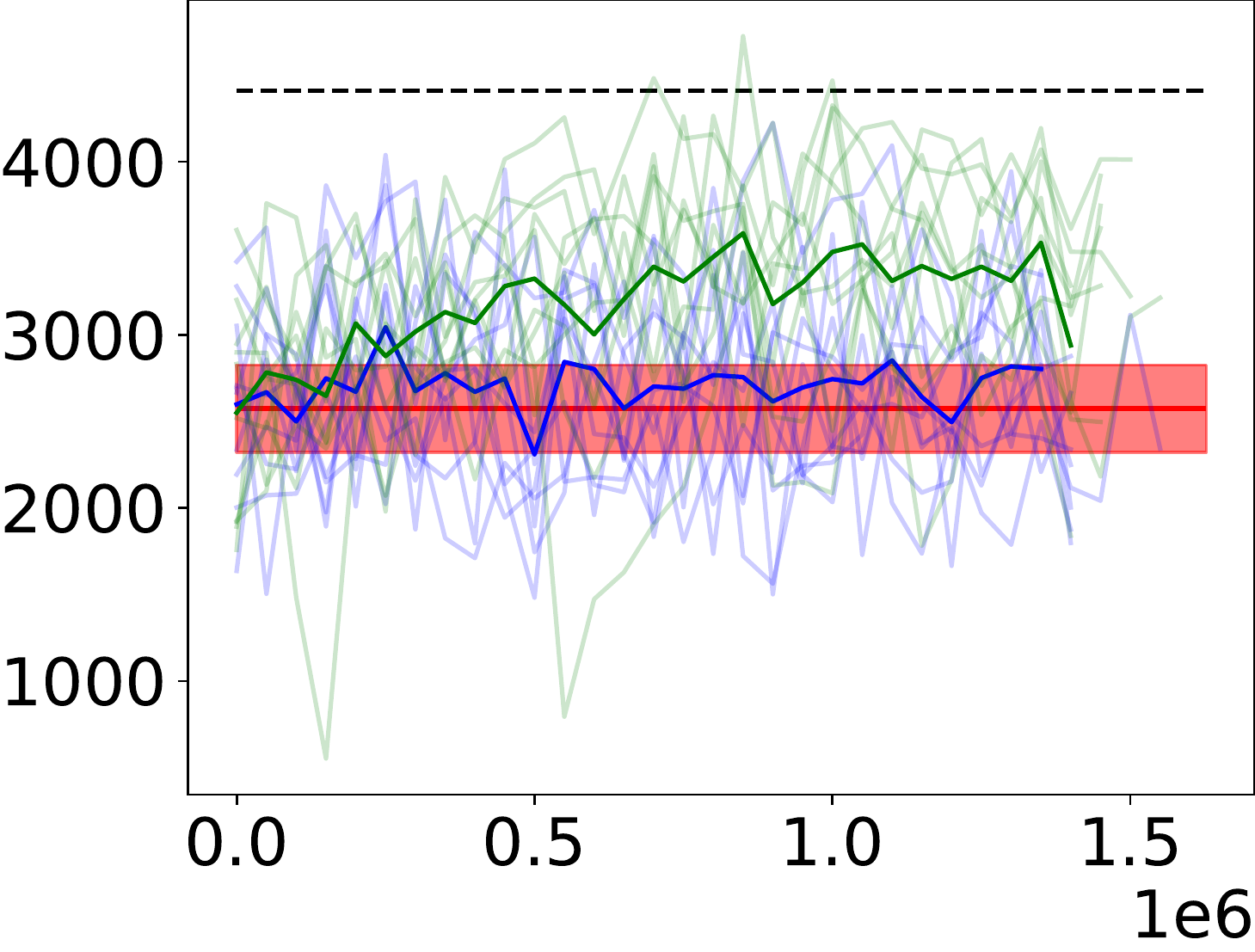}
	\end{subfigure}
	\begin{subfigure}{.24\textwidth}
		\centering
		\includegraphics[height=70pt]{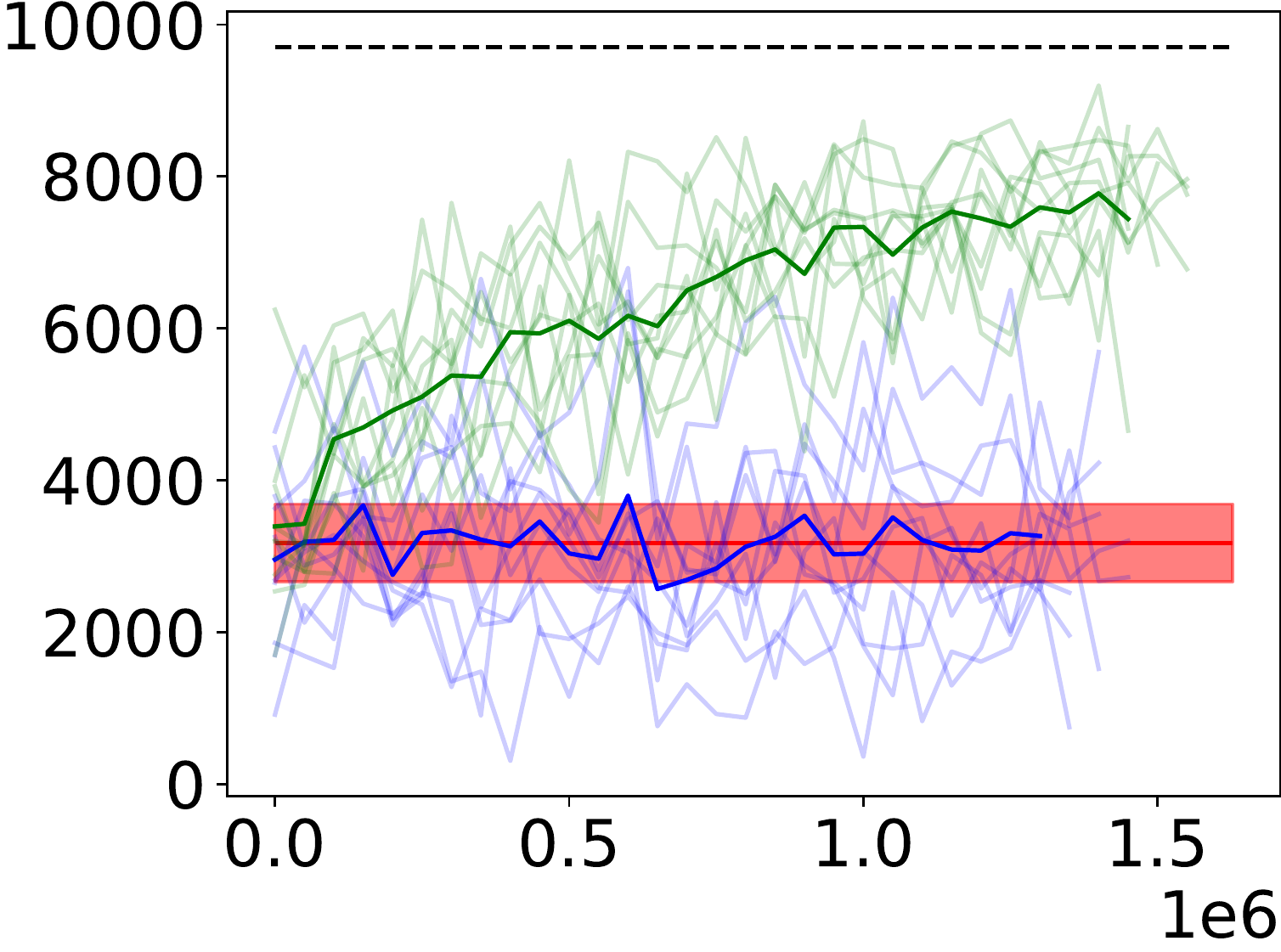}
	\end{subfigure}
	\begin{subfigure}{.24\textwidth}
		\centering
		\includegraphics[height=70pt]{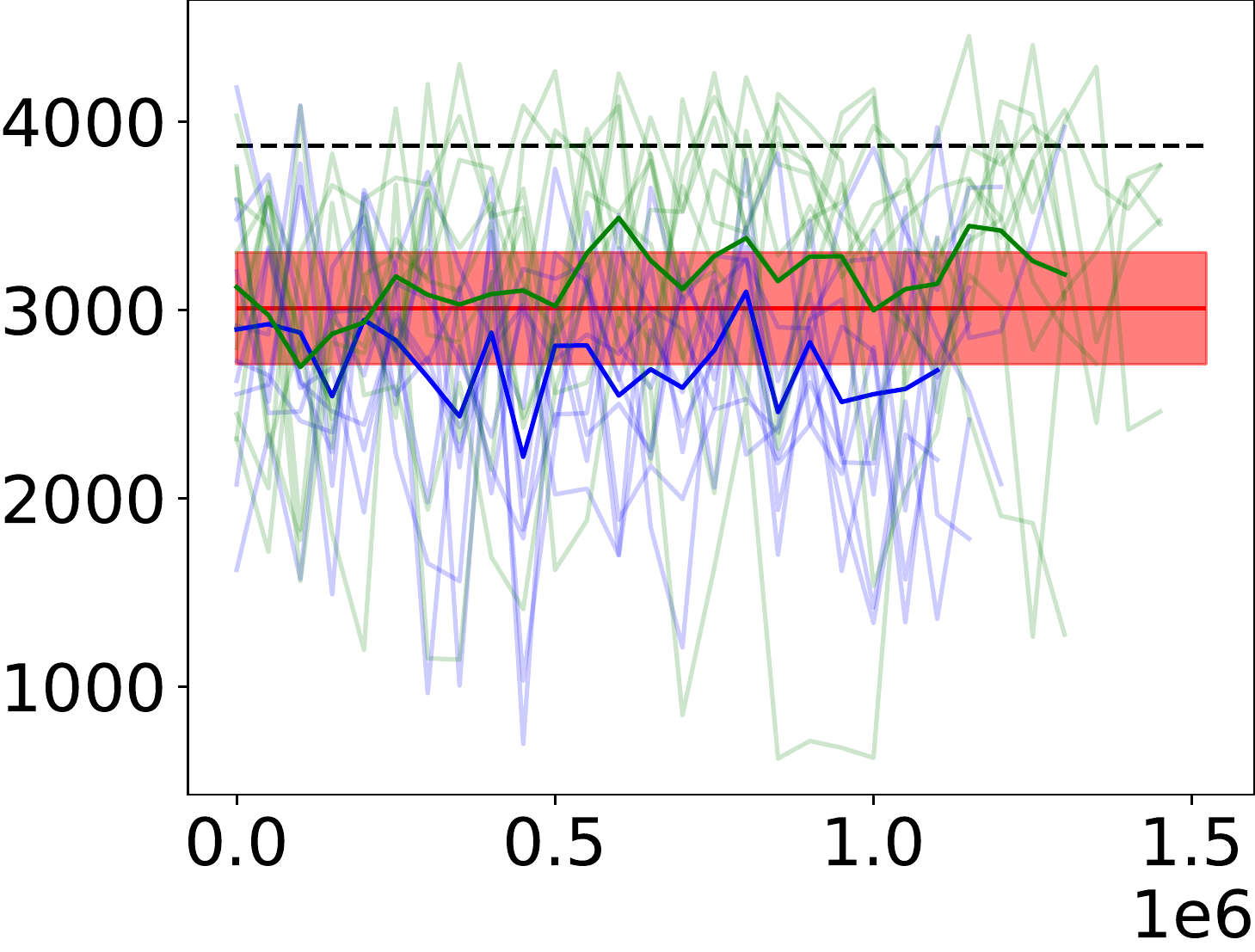}
	\end{subfigure}
	\begin{subfigure}{.24\textwidth}
		\centering
		\includegraphics[height=70pt]{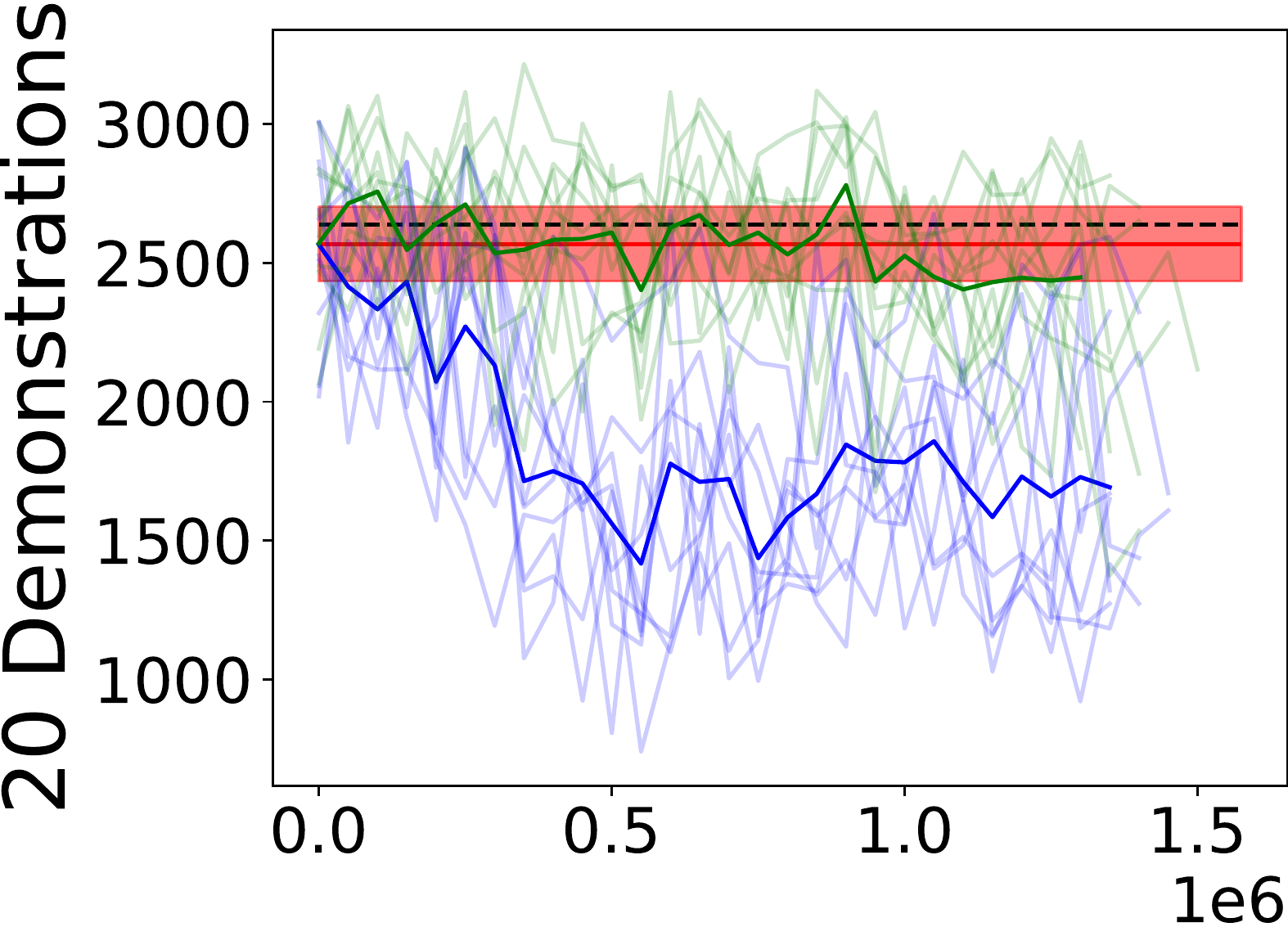}
	\end{subfigure}
	\begin{subfigure}{.24\textwidth}
		\centering
		\includegraphics[height=70pt]{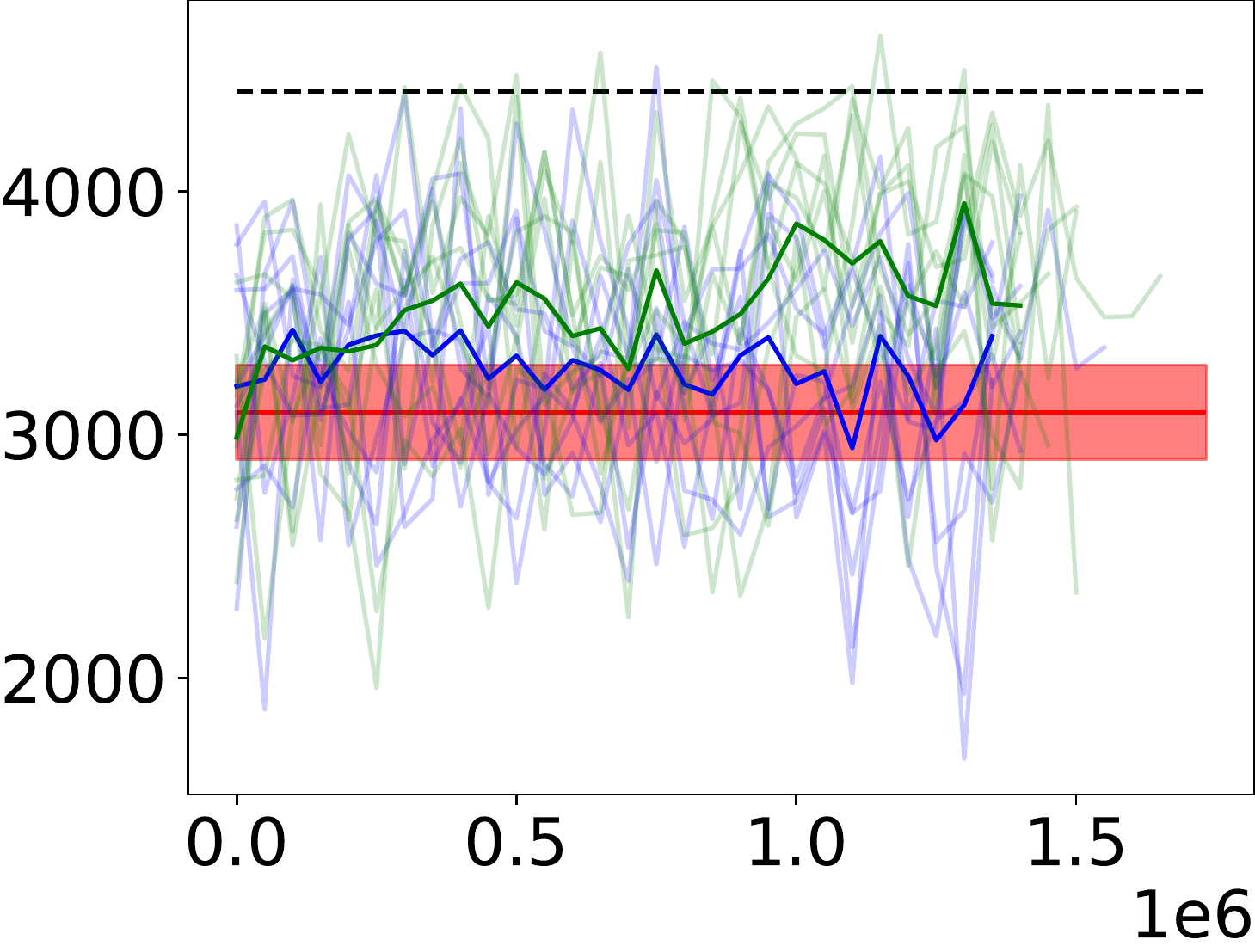}
	\end{subfigure}
	\begin{subfigure}{.24\textwidth}
		\centering
		\includegraphics[height=70pt]{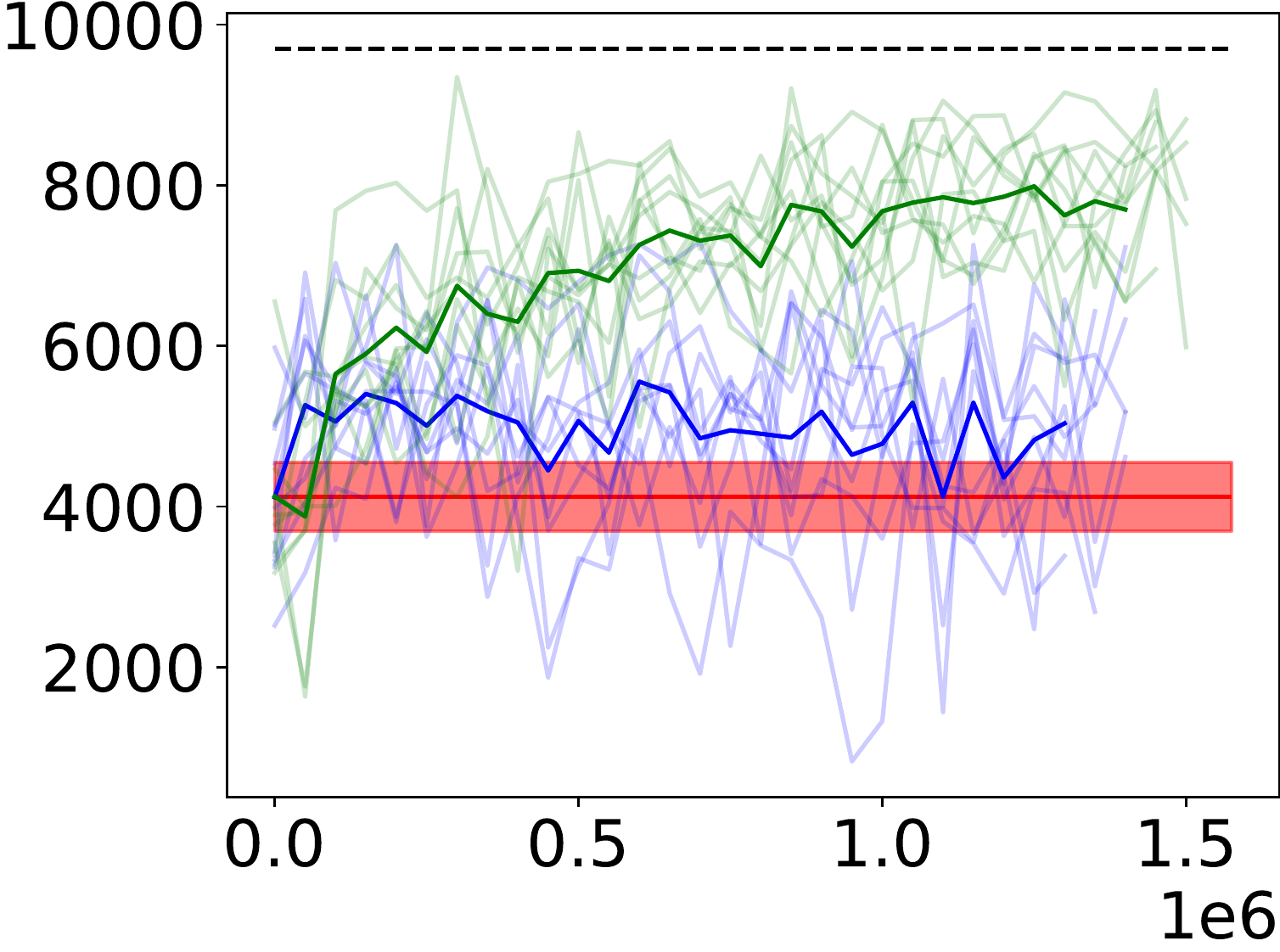}
	\end{subfigure}
	\begin{subfigure}{.24\textwidth}
		\centering
		\includegraphics[height=70pt]{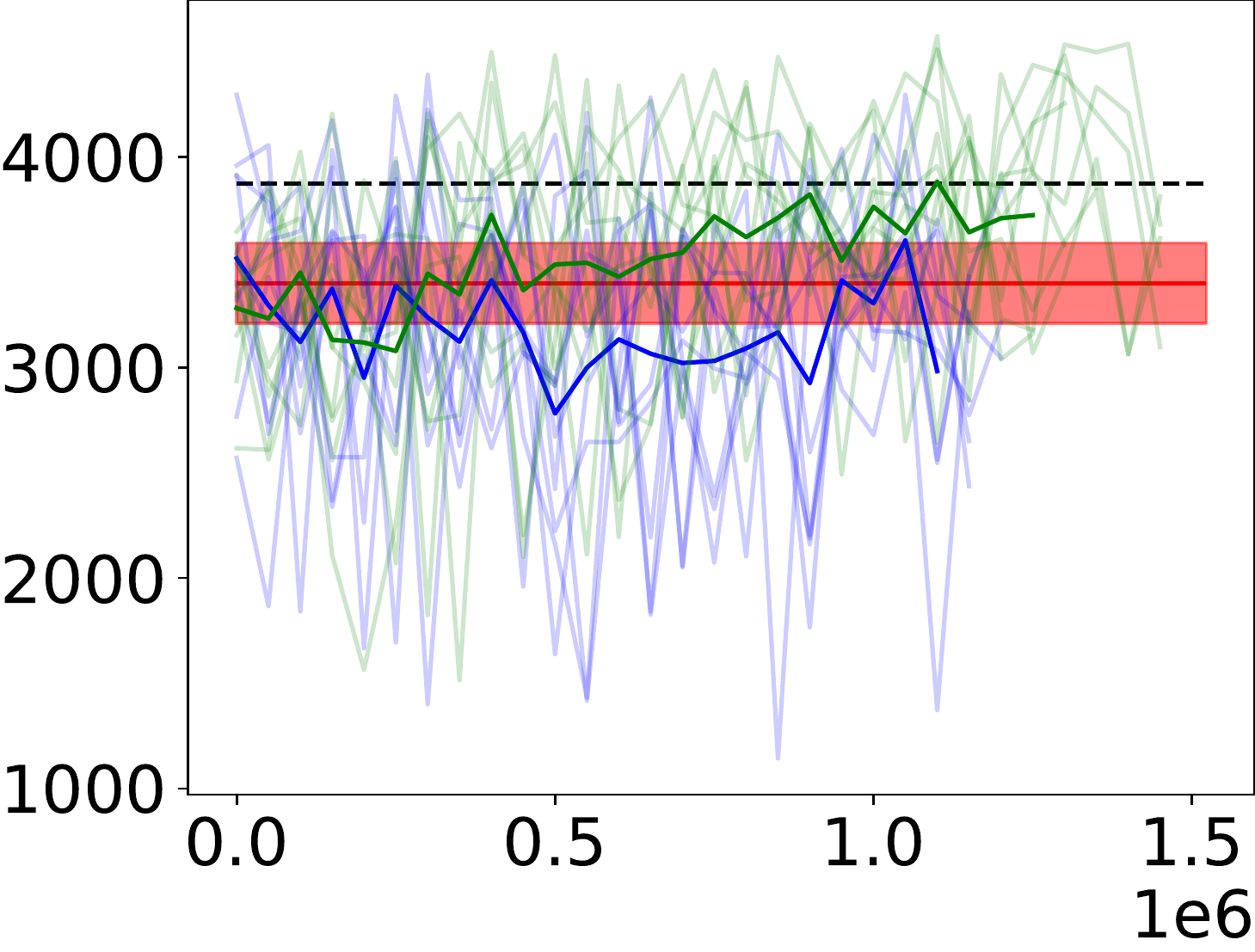}
	\end{subfigure}
	\caption{The plots show the approximated return (average of ten roll-outs) of the stochastic policy for ten trials for \gls{ONAIL} and ValueDice for different numbers of demonstrations and environments. For behavioral cloning, the shaded region  corresponds to the 0.95 confidence region based on the empirical standard error of the mean (when Gaussianity is assumed).}
	\label{nail_fig:halfcheetah_experiments}
\end{figure}

We compared the policy updates of \gls{ONAIL} and ValueDice on the Mujoco~\citep{Todorov2012} \textit{Hopper}, \textit{Walker2d}, \textit{HalfCheetah} and \textit{Ant} experiment. We obtained 50 expert demonstrations---each corresponding to one trajectory consisting of thousand steps---from a policy that was trained using soft actor-critic \citep[\gls{SAC},][]{Haarnoja2018}. We initialized the policy using behavioral cloning. More precisely, we used ten per cent of the training data as validation data for early-stopping, and trained the policy by maximizing the likelihood of the remaining expert data. Both algorithms are evaluated based on the same implementation\footnote{The code can be downloaded from \url{https://www.github.com/OlegArenz/O-NAIL}.} and differ only due to the different policy loss and the chosen hyper-parameters. For ValueDice, the policy updates are performed based on a mini-batch approximation of the saddle point problem given by Equation~\ref{nail_eq:VD_obj}. For \gls{ONAIL}, we use a mini-batch approximation of the \gls{iprojection}-loss (Eq.~\ref{nail_eq:nail_actor_loss}). Both algorithms use a mini-batch size of \num{256}. The Q-function and policy are each represented by a neural networks with two hidden layers of \num{256} units. The policy network outputs the parameters of a Gaussian distribution (with a diagonal state-dependent covariance matrix) and the sampled action are squashed by a $\tanh$ to respect action bounds, as discussed by~\citet{Haarnoja2018}.

We tuned the learning rates $\eta_\pi$ and $\eta_Q$ for the policy and Q-function as well as the number of gradient steps $N_\pi$ and $N_Q$ for the policy optimization and Q-optimization. We performed a grid-search on these hyper-parameters and present the chosen hyper-parameters in Table~\ref{nail_tab:hypertable}.  
\begin{table}
	\begin{center}
		\begin{small}
			\begin{sc}
				\begin{tabular}{lcc}
					\toprule
					hyper-parameter & value (ValueDice) & value (\gls{ONAIL}) \\
					\midrule
					$\eta_\pi$ & \num{1e-5} & \num{1e-4} \\
					$\eta_Q$ & \num{1e-3} & \num{1e-3}  \\
					$N_\pi$ & \num{1} & \num{10000}  \\
					$N_Q$ & \num{5} & \num{1000}  \\
					\bottomrule
				\end{tabular}
			\end{sc}
		\end{small}
	\end{center}
	\caption{The table shows the hyper-parameters used by \gls{ONAIL} and ValueDice for the \textit{HalfCheetah} experiment. \gls{ONAIL} performed better for large number of gradient steps, whereas ValueDice performed best for few gradient updates before switching policy updates and Q-function updates.}
	\label{nail_tab:hypertable}
\end{table}

The results of our experiments are shown in Figure~\ref{nail_fig:halfcheetah_experiments}. Although \gls{ONAIL} seems to clearly outperform ValueDice on these experiment, we want to stress, that the purpose of our evaluation was not to show practical advantages, but rather to demonstrate that we can derive novel algorithms based on our non-adversarial formulation that achieve similar performance compared to adversarial methods, while working at a significantly different modus operandi---namely, by using three to four orders of magnitude more update step for the Q-function and policy. Regarding the performance in practice, we want to point out the following shortcomings of our evaluations:
\begin{itemize}
	\item \textbf{Lack of regularization.} We did not perform any regularization on the Q-function or policy. It is well-known that adversarial methods sometimes can significantly benefit from regularizing the discriminator (or Q-function for ValueDice). For example, ValueDice applied a gradient penalty similar to the one that was introduced for Wasserstein-\glspl{GAN}~\citep{Gulrajani2017} on the Q-function. We actually tried this gradient penalty also on our experiment and were not able to show a benefit. Still, although \citet{Kostrikov2020} focused on the online setting, they also performed experiments in the offline setting, where they could show that ValueDice can outperform behavioral cloning. We believe that by introducing some type of regularization, both algorithms could potentially also achieve significantly better performance on our implementation. However, since these improvements are often achieved by introducing inductive biases, which can widen the gap between theory and practice, we have limited the evaluation to a simple implementation of the respective algorithms.
	\item \textbf{Initialization. } We initialized the policies using behavioral cloning. When performing the same experiment with randomly initialized policies, ValueDice would perform similarly, while \gls{ONAIL} would fail to learn at all with the chosen hyper-parameters. The lack of convergence of~\gls{ONAIL} is to be expected because using many discriminator updates to estimate a density-ratio between distributions with non-overlapping support typically results in sharp decision-boundaries that do not possess meaningful gradients or maxima. However, when performing only few Q-function-updates between optimizing the policy---which corresponds to a strong type of early-stopping---, such problems can be mitigated. Rather than relying on strong regularization that further disconnect theory and practice, we applied behavioral cloning to ensure overlapping support, which also seems reasonable and applicable in practice.
\end{itemize}

\section{Discussion}
\label{nail_sec:discussion}
Many modern methods in imitation learning and inverse reinforcement learning are based on an adversarial formulation. These methods frame the problem of distribution-matching as a minimax game between a policy and a discriminator, and rely on small policy updates for showing convergence to a Nash equilibrium. In contrast to these methods, we formulate distribution-matching as an iterative lower-bound optimization by alternating between maximizing and tightening a bound on the reverse \gls{KL} divergence. This non-adversarial formulation enables us to drop the requirement of ``sufficiently small'' policy updates for proving convergence. Algorithmically, our non-adversarial formulation is very similar to previous adversarial formulations and differs only due to an additional reward term that penalizes deviations from the previous policy. 

\subsection{Limitations and Future Work}
As the resulting algorithms are very similar to their adversarial counterparts, it can be difficult to show significant differences in practice. Hence, in this work, we focused on the insights gained from the non-adversarial formulation. For example, we showed that adversarial inverse reinforcement learning, which was previously not well understood, can be straightforwardly derived from our non-adversarial formulation. However, eventually we would like to derive stronger practical advantages from our formulation. We demonstrated that the non-adversarial formulation can be used to derive novel algorithms by presenting \gls{ONAIL}, an actor-critic based offline imitation learning method and our comparisons with ValueDice suggest that the non-adversarial formulation may indeed be beneficial. However, we hope to further distinguish \gls{ONAIL} from prior work by building on the close connection between non-adversarial imitation learning and inverse reinforcement learning in order to learn generalizable reward functions offline.

Adversarial methods have been suggested for a variety of different divergences, including~\citep{Ghasemipour2020} but not limited to~\citep{Xiao2019} the family of $f$-divergences. The non-adversarial formulation is currently limited to the reverse \gls{KL} divergence and penalizes deviations from the previous policy based on the reverse \gls{KL} divergence. It is an open question, whether our lower bound can be generalized to other divergences, for example, when penalizing deviations based on different divergences.

\section*{Acknowledgements}
Calculations for this research were conducted on the Lichtenberg high performance computer of the TU Darmstadt.

\newpage
\renewcommand{\theHsection}{A\arabic{section}}
\appendix
\section{BCE-loss of the AIRL Discriminator}
\label{nail_app:bce_loss}
The binary cross entropy loss for the \gls{AIRL} discriminator is given by
\begin{align*}
J_\text{BCE}(\boldsymbol{\theta}) =& E_{\stateN,\actionN \sim q(\stateN,\actionN)} \left[ \log \left( \frac{
	\exp{ \Big(\bar{\nu}_{\boldsymbol{\theta}}(\stateN,\actionN) \Big)}}{ \pi(\actionN|\stateN) + \exp{ \Big(\bar{\nu}_{\boldsymbol{\theta}}(\stateN,\actionN)  \Big)}} \right) \right] \\
&+ E_{\stateN,\actionN \sim p^{\pi}(\stateN,\actionN)} \left[ \log \left( \frac{\pi(\actionN|\stateN)}{\pi(\actionN|\stateN) + \exp{ \Big(\bar{\nu}_{\boldsymbol{\theta}}(\stateN,\actionN) \Big)}} \right) \right] \\
=& E_{\stateN,\actionN \sim q(\stateN,\actionN)} \left[ 
\bar{\nu}_{\boldsymbol{\theta}}(\stateN,\actionN)  \right]
+ E_{\stateN,\actionN \sim p^{\pi}(\stateN,\actionN)} \left[ \log \pi(\actionN|\stateN) \right] \\
&- 2 E_{\stateN,\actionN \sim \mu(\stateN,\actionN)} \left[ \log \left( \pi(\actionN|\stateN) + \exp{ \Big(\bar{\nu}_{\boldsymbol{\theta}}(\stateN,\actionN) \Big)} \right) \right],
\end{align*}
where $\mu(\stateN, \actionN) = \frac{1}{2} (q(\stateN,\actionN) + p^\pi(\stateN,\actionN))$ is a mixture of the distributions induced by the expert and the agent. 
The gradient with respect to the discriminator parameters is given by
\begin{align}
\label{nail_eq:bce_gradient}
\frac{dJ_\text{BCE}(\boldsymbol{\theta})}{d\boldsymbol{\theta}} =&
E_{\stateN,\actionN \sim q(\stateN,\actionN)} \left[ 
\frac{d \bar{\nu}_{\boldsymbol{\theta}}(\stateN,\actionN)}{d \boldsymbol{\theta}}  \right]
- 2 E_{\stateN,\actionN \sim \mu(\stateN,\actionN)} \left[ \frac{\exp{ \Big(\bar{\nu}_{\boldsymbol{\theta}}(\stateN,\actionN) \Big)}}{\pi(\actionN|\stateN) + \exp{ \Big(\bar{\nu}_{\boldsymbol{\theta}}(\stateN,\actionN) \Big)}}
\frac{d \bar{\nu}_{\boldsymbol{\theta}}(\stateN,\actionN)}{d \boldsymbol{\theta}}
\right] \nonumber \\
=&
E_{\stateN,\actionN \sim q(\stateN,\actionN)} \left[ 
\frac{d \bar{\nu}_{\boldsymbol{\theta}}(\stateN,\actionN)}{d \boldsymbol{\theta}}  \right]
- E_{\stateN,\actionN \sim \mu(\stateN,\actionN)} \left[ \frac{{\bar{p}}_{\boldsymbol{\theta}}(\stateN,\actionN)}{\frac{1}{2} \left(
	p^\pi(\stateN,\actionN) + {\bar{p}}_{\boldsymbol{\theta}}(\stateN,\actionN) \right)}
\frac{d \bar{\nu}_{\boldsymbol{\theta}}(\stateN,\actionN)}{d \boldsymbol{\theta}}
\right] \nonumber \\
=&
E_{\stateN,\actionN \sim q(\stateN,\actionN)} \left[ 
\frac{d \bar{\nu}_{\boldsymbol{\theta}}(\stateN,\actionN)}{d \boldsymbol{\theta}}  \right]
- E_{\stateN,\actionN \sim \mu(\stateN,\actionN)} \left[ \frac{{\bar{p}}_{\boldsymbol{\theta}}(\stateN,\actionN)}{\bar{\mu}(\stateN,\actionN)}
\frac{d \bar{\nu}_{\boldsymbol{\theta}}(\stateN,\actionN)}{d \boldsymbol{\theta}}
\right],
\end{align}
where we introduced ${\bar{p}}_{\boldsymbol{\theta}}(\stateN,\actionN) = p^\pi(\stateN) \exp\left( \bar{\nu}_{\boldsymbol{\theta}}(\stateN,\actionN) \right)$ and ${\bar{\mu}}(\stateN,\actionN)=\frac{1}{2} \left(
p^\pi(\stateN,\actionN) + {\bar{p}}_{\boldsymbol{\theta}}(\stateN,\actionN) \right)$.

\citet{Fu2018} argue that, when assuming that the policy $\pi$ maximizes the policy objective, we would have $\bar{p}_{\boldsymbol{\theta}}(\stateN, \actionN) = p_{\boldsymbol{\theta}}(\stateN, \actionN)$ and that the gradient (Eq.~\ref{nail_eq:bce_gradient}) would then correspond to an importance-sampling based estimate of the maximum-likelihood gradient (Eq.~\ref{nail_eq:maxent_gradient}).
However, for obtaining the correct importance weights, we would need to further assume that $ \bar{p}_{\boldsymbol{\theta}}(\stateN, \actionN) = q(\stateN, \actionN)$ such that  $\bar{\mu}(\stateN,\actionN)=\mu(\stateN,\actionN)$, that is, we would need to assume that the distribution induced by the current policy $p^{\pi}(\stateN, \actionN) = \bar{p}_{\boldsymbol{\theta}}(\stateN, \actionN)$ matches the expert distribution. Furthermore, we would additionally need to assume that the discriminator is optimal such that $\bar{\nu}_{\boldsymbol{\theta}}(\stateN,\actionN) = A^{\text{soft},\pi}(\stateN,\actionN) = \log{\pi(\actionN|\stateN)}$ in order to ensure $\bar{p}_{\boldsymbol{\theta}}(\stateN, \actionN) = p_{\boldsymbol{\theta}}(\stateN, \actionN)$. While it is reassuring that both methods share a stationary point when the expert distribution is perfectly matched and the policy and discriminator are optimal, the connection between \gls{AIRL} and \gls{MaxCausalEntIRL} presented by~\citet{Fu2018} seems rather weak.


\section{Proof for Proposition~\ref{nail_prop:stateReward}}
\label{nail_app:proof_markovian_reward}
	Let 
	\begin{equation*}
	p(t|\traj) = 
	\begin{cases}
	\frac{1}{T(\traj)} & t < T(\traj) \\
	0 & otherwise
	\end{cases}
	\end{equation*}
	denote the probability of observing the time step $t$ when the trajectory
	$\traj$ of length $T(\traj)$ is given.
Based on the assumption $p(\obN|\traj,t) = p(\obN|\state{t}^{\traj}, \action{t}^{\traj})$, we can express $p(\obN|\traj)$ as follows:
\begin{equation}
\label{nail_eq:prop_state_reward_eq1}
p(\obN|\traj) = \sum_{t=0}^{\infty} p(t|\traj) p(\obN|\traj,t) = \sum_{t=0}^{\infty} p(t|\traj) p(\obN|\state{t}^{\traj}, \action{t}^{\traj}).
\end{equation}
Based on equation~\ref{nail_eq:prop_state_reward_eq1}, we can express the objective of the episodic reinforcement learning problem given by Eq.~\ref{nail_eq:episodic_rl} as
\begin{align}
\label{nail_eq:adv_step}
J_\text{rl,ep}(\pi) &= \mathrm{E}_{\traj \sim p^\pi(\traj)}\left[  \int_\obN p(\obN|\traj) f^*(D(\obN)) d\obN \right] \nonumber \\
&=  \int_{\traj} p^{\pi}(\traj)  \int_\obN  \sum_{t=0}^{\infty} p(t|\traj) p(\obN|\state{t}^{\traj}, \action{t}^{\traj}) f^*(D(\obN)) d\obN d\traj \nonumber \\
&= \sum_{t=0}^{\infty} p(t) \int_{\traj}  p^{\pi}(\traj|t)    \int_\obN p(\obN|\state{t}^{\traj}, \action{t}^{\traj}) f^*(D(\obN)) d\obN d\traj \nonumber \\
&= \sum_{t=0}^{\infty}  p(t)  \int_{\stateN, \actionN} p_{t}^{\pi}(\stateN, \actionN)  \int_\obN  p(\obN|\stateN, \actionN) f^*(D(\obN)) d\obN d\stateN d\actionN \nonumber \\
&= (1 - \gamma) \sum_{t=0}^{\infty} \gamma^t \int_{\stateN, \actionN}  p_{t}^{\pi}(\stateN, \actionN)  \int_\obN  p(\obN|\stateN, \actionN) f^*(D(\obN)) d\obN d\stateN d\actionN \nonumber \\
&= (1 - \gamma) \sum_{t=0}^{\infty} \gamma^t \int_{\stateN, \actionN}  p_{t}^{\pi}(\stateN, \actionN)  r_{\text{adv}}(\stateN, \actionN) d\stateN d\actionN  \nonumber\\
&= \int_{\stateN, \actionN} p^{\pi}(\stateN, \actionN) r_{\text{adv}}(\stateN, \actionN) d\stateN d\actionN.
\end{align}
Hence, we can solve the episodic reinforcement learning problem (Eq.~\ref{nail_eq:episodic_rl}) also by maximizing the expected Markovian reward $r_{\text{adv}}(\stateN, \actionN) = \int_\obN  p(\obN|\stateN, \actionN) f^*(D(\obN)) d\obN$. When defining $p(\obN|\stateN, \actionN)$ as a delta distribution at $\obN(\stateN, \actionN) = [\stateN^\top, \actionN^\top ]^\top$ or at $\obN(\stateN, \actionN) = \stateN$ we can also recover the common objective of matching the expert's state-action distribution or its state marginal. However, such restrictions are not necessary to obtain Markovian rewards. \qed 

\section{Proof for Lemma~\ref{NAIL_LEMMA:LOWERBOUND}}
\label{nail_app:proof_nail_lb}
Based on Equation~\ref{nail_eq:lowerbound_plus_EKL}, we can express the reverse KL divergence $D_\text{RKL}(\tilde{p}(\stateN, \actionN)||q(\stateN, \actionN))$ in terms of the lower bound $J_{\text{NAIL},\tilde{\pi}}(\tilde{\pi})$ (Eq.\ref{nail_eq:lb_objective}) as 
\begin{align*}
D_\text{RKL}(\tilde{p}(\stateN, \actionN)||q(\stateN, \actionN)) = -J_{\text{NAIL},\tilde{\pi}}(\tilde{\pi}) - \mathrm{E}_{\tilde{p}(\mathbf{o})} \left[ \text{KL}\left(\tilde{p}(\boldsymbol{\tau}|\mathbf{o})||\tilde{p}(\boldsymbol{\tau}|\mathbf{o})\right) \right] 
= -J_{\text{NAIL},\tilde{\pi}}(\tilde{\pi}),
\end{align*}
and the reverse KL for a given policy $\pi$ as
\begin{align*}
D_\text{RKL}(p^{\pi}(\stateN, \actionN)||q(\stateN, \actionN)) = -J_{\text{NAIL},\tilde{\pi}}(\pi) - \mathrm{E}_{p^\pi(\mathbf{o})} \left[ \text{KL}\left(p^{\pi}(\boldsymbol{\tau}|\mathbf{o})||\tilde{p}(\boldsymbol{\tau}|\mathbf{o})\right) \right].
\end{align*}

Hence, for any policy $\pi$ that satisfies $J_{\text{NAIL},\tilde{\pi}}(\pi) > J_{\text{NAIL},\tilde{\pi}}(\tilde{\pi})$, we have
\begin{align*}
&D_\text{RKL}(\tilde{p}(\stateN, \actionN)||q(\stateN, \actionN)) - D_\text{RKL}(p^{\pi}(\stateN, \actionN)||q(\stateN, \actionN)) \\ &\quad=  \underbrace{J_{\text{NAIL},\tilde{\pi}}(\pi) - J_{\text{NAIL},\tilde{\pi}}(\tilde{\pi})}_{> 0} + \underbrace{\mathrm{E}_{p^\pi(\mathbf{o})} \left[ \text{KL}\left(p^{\pi}(\boldsymbol{\tau}|\mathbf{o})||\tilde{p}(\boldsymbol{\tau}|\mathbf{o})\right) \right]}_{>0} > 0.
\end{align*}
\qed

\section{Proof for Theorem~\ref{NAIL_THM:NAIL_ALG}}
\label{nail_app:proof_nail_alg}
The sequence $\left\{D_\text{RKL}(p^{\pi^{(i)}}(\stateN, \actionN)||q(\stateN, \actionN))\right\}_{i=0}^{\infty}$ is monotonously decreasing (see Lemma~\ref{NAIL_LEMMA:LOWERBOUND}) and bounded below (due to the non-negativity of the KL) and thus convergent~\citep{Bibby1974}. At convergence $\pi^{(i)}$ must be a stationary point of the lower bound objective (otherwise we could improve using gradient descent), that is
\begin{align*}
0 = \frac{d}{d\pi} J_{\text{NAIL},\pi^{(i)}}(\pi)\bigg\rvert_{\pi=\pi^{(i)}} =& \frac{d}{d\pi} D_\text{RKL}(p^{\pi}(\stateN, \actionN)||q(\stateN, \actionN))\bigg\rvert_{\pi=\pi^{(i)}}  \\ &+ \underbrace{\frac{d}{d\pi} \mathrm{E}_{p^\pi(\mathbf{o})} \left[ \text{KL}\left(p^{\pi}(\boldsymbol{\tau}|\mathbf{o})||p^{\pi^{(i)}}(\boldsymbol{\tau}|\mathbf{o})\right) \right]\bigg\rvert_{\pi=\pi^{(i)}} }_{=0}.
\end{align*}
Hence, $\pi^{(i)}$ is a stationary point of the KL objective, that is,
\begin{align*}
\frac{d}{d\pi} D_\text{RKL}(p^{\pi}(\stateN, \actionN)||q(\stateN, \actionN))\bigg\rvert_{\pi=\pi^{(i)}} = 0.
\end{align*}
\qed

\section{Proof for Lemma~\ref{NAIL_LEMMA:Q_RELATIONS}}
\label{nail_app:proof_Q_relations}
\citet{Haarnoja2018} showed that the soft Q-function for policy $\pi$ and reward $r$ can be learned by repeatably applying the modified Bellman backup operator
\begin{equation}
\label{nail_eq:softpolicyeval}
\mathcal{T}^\pi Q^{\text{soft}}(\state{t},\action{t}) \triangleq r(\state{t}, \action{t}) + \gamma \mathrm{E}_{\state{t+1} \sim p} \left[ V^{\text{soft}}(\state{t+1}) \right]
\end{equation}
where 
\begin{equation}
V^\text{soft}(\stateN) = \mathrm{E}_{\actionN \sim \pi} \left[ Q^{\text{soft}}(\stateN,\actionN) - \log \pi(\actionN|\stateN) \right].
\end{equation}
We will now prove that $\hat{Q}(\stateN,\actionN) \triangleq Q_{r}^{\tilde{\pi}}(\stateN,\actionN) + \log \tilde{\pi}(\actionN|\stateN)$ is the soft Q-function for policy $\tilde{\pi}$ and lower bound reward $r_{\text{lb}}(\stateN, \actionN) = r(\stateN,\actionN) + \log \tilde{\pi}(\actionN|\stateN)$, that is $\hat{Q} = Q_{r_\text{lb}}^{\text{soft},\tilde{\pi}}$, by showing that it is a fixed point of the modified Bellman backup operator (Eq.~\ref{nail_eq:softpolicyeval}).

Applying the modified Bellman update to $\hat{Q}(\stateN,\actionN)$ yields
\begin{align*}
\mathcal{T}^{\tilde{\pi}} \hat{Q}(\state{t}, \action{t}) &=r_{\text{lb}}(\state{t}, \action{t}) + \gamma \mathrm{E}_{\state{t+1} \sim p, \action{t+1} \sim \tilde{\pi}} \left[ \hat{Q}(\stateN,\actionN)  - \log \tilde{\pi}(\action{t+1}|\state{t+1}) \right] \\
&= r(\state{t}, \action{t}) +  \log \tilde{\pi}(\action{t}|\state{t}) \\
&\phantom{=}+ \gamma \mathrm{E}_{\state{t+1} \sim p, \action{t+1} \sim \tilde{\pi}} \Big[ Q_{r}^{\tilde{\pi}}(\state{t+1},\action{t+1}) + \log \tilde{\pi}(\action{t+1}|\state{t+1}) - \log \tilde{\pi}(\action{t+1}|\state{t+1}) \Big] \\
&= r(\state{t}, \action{t}) +  \log \tilde{\pi}(\action{t}|\state{t}) + \gamma \mathrm{E}_{\state{t+1} \sim p, \action{t+1} \sim \tilde{\pi}} \left[ Q_{r}^{\tilde{\pi}}(\state{t+1}, \action{t+1}) \right]
\\
&= Q_{r}^{\tilde{\pi}}(\state{t},\action{t}) + \log \tilde{\pi}(\action{t}|\state{t}) = \hat{Q}(\stateN,\actionN).
\end{align*}
\qed

\section{Proof of Lemma~\ref{NAIL_LEMMA:NAIL_ACTOR_STEP}}
\label{nail_app:proof_actor_step}
The first part of the proof closely follows the proof given by~\citet{Haarnoja2018}, which itself closely follows the proof of the policy improvement theorem given by \citet{SuttonBarto1998}. However, \citet{Haarnoja2018} only considered policies that are optimal with respect to the \gls{iprojection}-loss (Eq.~\ref{nail_eq:nail_actor_loss}), although the proof is also valid for the less strict assumption of policy improvement, as shown below.

Based on our assumption we have for any state \stateN
\begin{align}
\begin{split}
\label{nail_eq:soft_actor_improvement}
\mathrm{E}_{\actionN \sim \pi(\actionN|\stateN)} \left[ Q_{r_\text{lb}}^{\text{soft},{\pi}^{(i)}}(\stateN, \actionN) - \log{\pi}(\actionN|\stateN)  \right] \ge &\mathrm{E}_{\actionN \sim \pi^{(i)}(\actionN|\stateN)} \left[ Q_{r_\text{lb}}^{\text{soft},{\pi}^{(i)}}(\stateN, \actionN) - \log{\pi^{(i)}}(\actionN|\stateN) \right] \\
&= V_{r_\text{lb}}^{\text{soft},{\pi}^{(i)}}(\stateN).
\end{split}
\end{align}

Based on Equation~\ref{nail_eq:soft_actor_improvement} we can repeatedly apply the soft Bellman equation to bound the soft Q-function for the old policy by the soft Q-function for the new policy, as follows:
\begin{align}
\begin{split}
\label{nail_eq:newQlargeroldQ}
Q_{r_\text{lb}}^{\text{soft},{\pi}^{(i)}}(\state{t}, \action{t}) &= r_\text{lb}(\state{t}, \action{t}) + \gamma \mathrm{E}_{\state{t+1} \sim p} \left[ V^{\text{soft}}(\state{t+1}) \right] \\
&\le r_\text{lb}(\state{t}, \action{t}) + \gamma \mathrm{E}_{\state{t+1} \sim p, \action{t+1} \sim \pi} \left [ Q_{r_\text{lb}}^{\text{soft},{\pi}^{(i)}}(\state{t+1}, \action{t+1}) - \log{\pi}(\action{t+1}|\state{t+1})  \right] \\
&\; \;\vdots \\
&\le Q_{r_\text{lb}}^{\text{soft},{\pi}}(\state{t}, \action{t}).
\end{split}
\end{align}

We now express the lower bound objective $J_{\text{NAIL}, \pi^{(i)}}(\pi)$ in terms of its Q-function $Q_{r_\text{lb}}^{\text{soft},{\pi}}$ and relate it to the lower bound objective of the last policy $J_{\text{NAIL}, \pi^{(i)}}(\pi^{(i)})$ using the inequalities \ref{nail_eq:newQlargeroldQ} and \ref{nail_eq:soft_actor_improvement}, namely,
\begin{align*}
J_{\text{NAIL}, \pi^{(i)}}(\pi) =& \int_{\stateN,\actionN} p^\pi(\stateN,\actionN) \left( r_\text{lb}^{\pi^{(i)}}(\stateN, \actionN) - \log \pi(\actionN|\stateN) \right) d\stateN d\actionN \\
=& (1-\gamma) \mathrm{E}_{\stateN \sim p_{0}(\stateN), \actionN \sim \pi} \left[ Q_{r_\text{lb}^{\pi^{(i)}}}^{\text{soft},{\pi}} - \log \pi(\actionN|\stateN) \right] 
	\\
\ge& (1-\gamma) \mathrm{E}_{\stateN \sim p_{0}(\stateN), \actionN \sim \pi} \left[ Q_{r_\text{lb}^{\pi^{(i)}}}^{\text{soft},{\pi^{(i)}}} - \log \pi(\actionN|\stateN) \right] 
\\
\ge&
(1-\gamma) \mathrm{E}_{\stateN \sim p_{0}(\stateN), \actionN \sim \pi^{(i)}} \left[ Q_{r_\text{lb}^{\pi^{(i)}}}^{\text{soft},{\pi^{(i)}}} - \log \pi^{(i)}(\actionN|\stateN) \right] 
= J_{\text{NAIL}, \pi^{(i)}}(\pi^{(i)}).
\end{align*}
\qed

\newpage
\bibliography{papers}

\end{document}